\documentclass[10pt,twocolumn,letterpaper]{article}

\usepackage{cvpr}
\usepackage{times}
\usepackage{epsfig}
\usepackage{graphicx}
\usepackage{amsmath}
\usepackage{amssymb}
\usepackage{microtype}
\usepackage{graphicx}
\usepackage{subfigure}
\usepackage{booktabs} 

\usepackage{amsmath,amsthm}
\usepackage{amsfonts}

\usepackage[breaklinks=true,bookmarks=false]{hyperref}

\cvprfinalcopy 

\def\cvprPaperID{****} 

\def\N{{\mathcal N}}
\def\L{{\mathcal L}}
\def\P{{\mathcal P}}
\def\G{{\mathcal G}}
\def\E{{\mathbb  E}}
\def\F{{\mathcal F}}

\def\C{{\mathcal C}}

\begin{document}

\title{A Deep Optimization Approach for Image Deconvolution}

\author{
Zhijian Luo$^{\dagger}$,Siyu Chen$^{\dagger}$\\
Zhejiang University\\
No. 38, ZheDa Road.\\
{\tt\small \{luozhijian,sychen\}@zju.edu.cn}\\
\and
Yuntao Qian\\
Zhejiang University\\
No. 38, ZheDa Road.\\
{\tt\small ytqian@zju.edu.cn}
}

\maketitle

\let\thefootnote\relax\footnotetext[1]{$\dagger$ equal contribution}

\begin{abstract} \label{sec:abs}
In blind image deconvolution, priors are often leveraged to constrain the solution space, so as to alleviate the under-determinacy.
Priors which are trained separately from the task of deconvolution tend to be instable, or ineffective.
We propose the Golf Optimizer, a novel but simple form of network that learns deep priors from data with better propagation behavior.
Like playing golf, our method first estimates an aggressive propagation towards optimum using one network, and recurrently applies a residual CNN to learn the gradient of prior for delicate correction on restoration.
Experiments show that our network achieves competitive performance on GoPro dataset, and our model is extremely lightweight compared with the state-of-art works.
\end{abstract}

\section{Introduction} \label{sec:intro}
Blind image deconvolution, which restores an unknown latent image from blurry degeneration,
is a fundamental task in image processing and computer vision.
The most commonly used formulation of blur degeneration $y$ is modeled as the convolution of the latent image $x$ and the kernel $k$:
\begin{equation}\label{eq:model}
  y = x \ast k + n,
\end{equation}
where $\ast$ denotes the convolution operator and $n$ is i.i.d Gaussian noise.
Blind image deconvolution aims to estimate the latent image $x$ given a blurry image $y$, and it is highly ill-posed since both $k$ and $n$ are unknown.
To tackle this problem, prior knowledge is required to constrain the solution space.
Recently, deep convolutional neural networks (CNNs) have been applied to image deconvolution and achieved significant improvements.
Due to its powerful approximation capability,  such networks can implicitly incorporates image prior information\cite{schuler2016learning, nah2017deep, tao2018srndeblur, noroozi2017motion}.
Besides, many CNN-based methods directly estimate sharp images with trainable networks which introduce explicit deep generative priors \cite{ramakrishnan2017deep,li2018learning,asim2018deep,DeblurGAN}.

\begin{figure}[t!]
\vskip 0.2in
\begin{center}
\centerline{\includegraphics[width=1.0\linewidth]{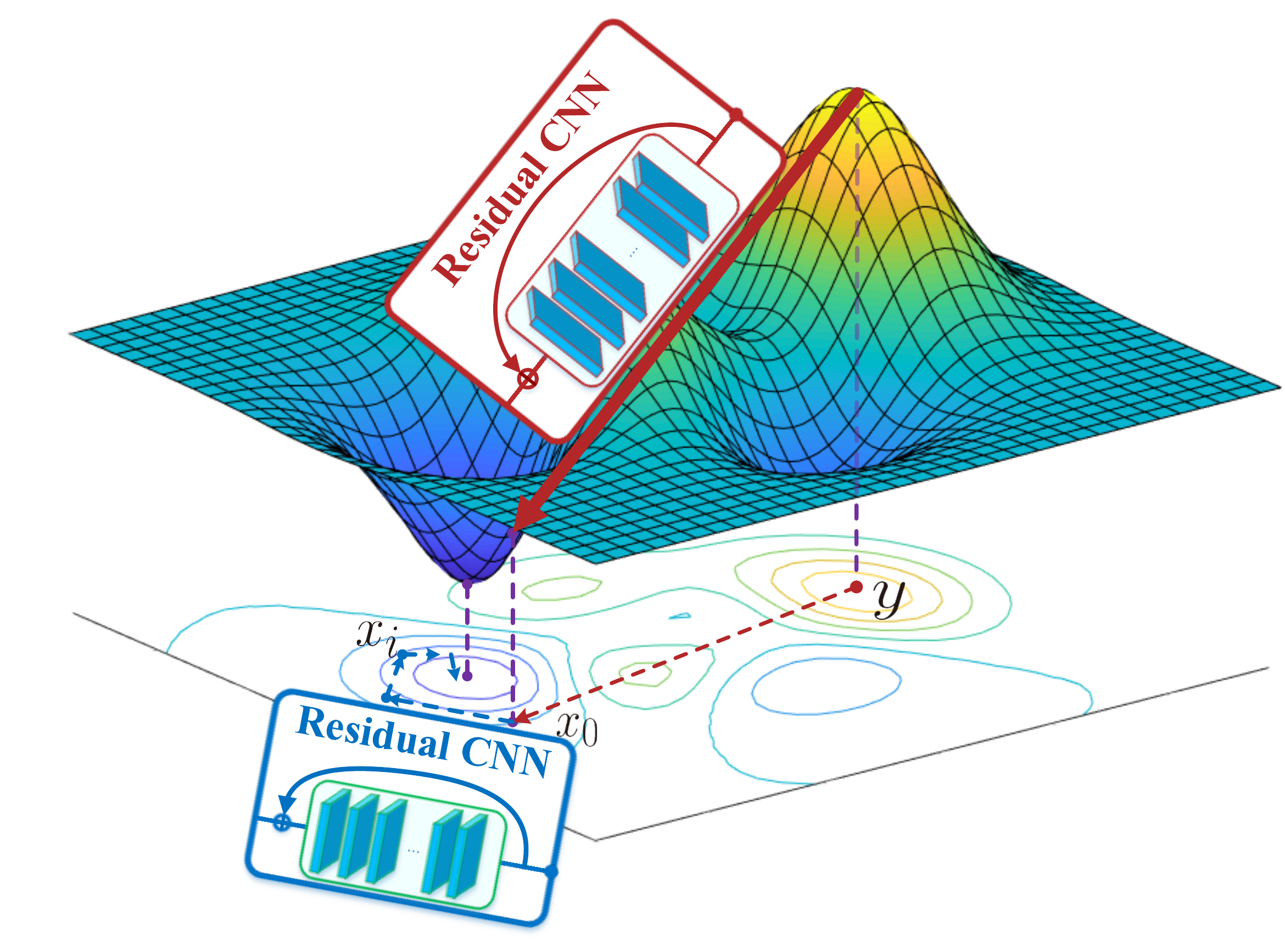}}\caption{The illustration of our framework. Red arrow denotes the aggressive propagation of first network shown in red block with Residual CNN, and blue dotted arrow denotes the delicate correction of second network shown in blue block.}\label{fig:flowchart}
\end{center}
\vskip -0.2in
\end{figure}

In this paper, we present a novel way for blind deconvolution with a general and tidy framework.
The proposed method learns data-driven priors using an optimizer, which is separated into two task-dependent networks.
Figure \ref{fig:flowchart} provides the illustration of our proposed framework.
The first network shown in red block tries to estimate an aggressive propagation towards the optimum by a vanilla residual CNN;
while the second network shown in blue block employs recurrent residual unit of ResNet \cite{he2016deep} and behaves like an iterative optimizer for delicate correction on the restored image.
The behavior of our framework is similar to an expert golf player, who tries to get the ball near to the hole at first shot.
Then, to get the ball close to, or even in the hole, the player taps the ball towards the optimum with delicate adjustments.
Hence, we refer to the optimizer as \textit{Golf Optimizer}.

It is obvious that trivially applying iterative optimizer for blind image deconvolution would lead to multi-tasking, since the image restored by the optimizer after every iteration would not still follow the physical model of blur degradation.
This changes the data distribution and forces the deconvolution optimizer to accommodate new knowledge, which could lead to \textit{multi-tasking} and cause \textit{catastrophic forgetting} \cite{Kirkpatrick3521,serra2018overcoming,lee2017overcoming}. 
We are aware of this phenomenon in the training of iterative optimizer network for blind deconvolution, and to the best of our knowledge it is the first time this phenomenon is addressed in image deconvolution.
To alleviate this phenomenon, we employ a vanilla residual CNN as our first part of network to preprocess images, in which the update is performed with an aggressive propagation.

The key insight of our optimizer lies on the asymptotical learning of the \textit{gradient of prior}.
Unlike previous works concentrated on learning image prior with a deconvolution irrelevant objective (e.g. classification error or denoising error) \cite{li2018learning,zhang2017learning}, the Golf Optimizer learns the image priors within the deconvolution task.
Furthermore, the largest challenge in prior learning is the instability of network when employing discriminative/qualitative criterion on restoration quality \cite{DeblurGAN,ramakrishnan2017deep,asim2018deep}.
In practice, these priors may be incapable of providing optimal information for deconvolution \cite{asim2018deep}.
To eliminate this instability, instead of learning image prior itself, our optimizer asymptotically learns the gradient of prior via training recurrent residual unit of ResNet \cite{he2016deep}.

\begin{figure}[t!]
\vskip 0.2in
\begin{center}
\subfigure[Input]{\includegraphics[width=0.32\linewidth]{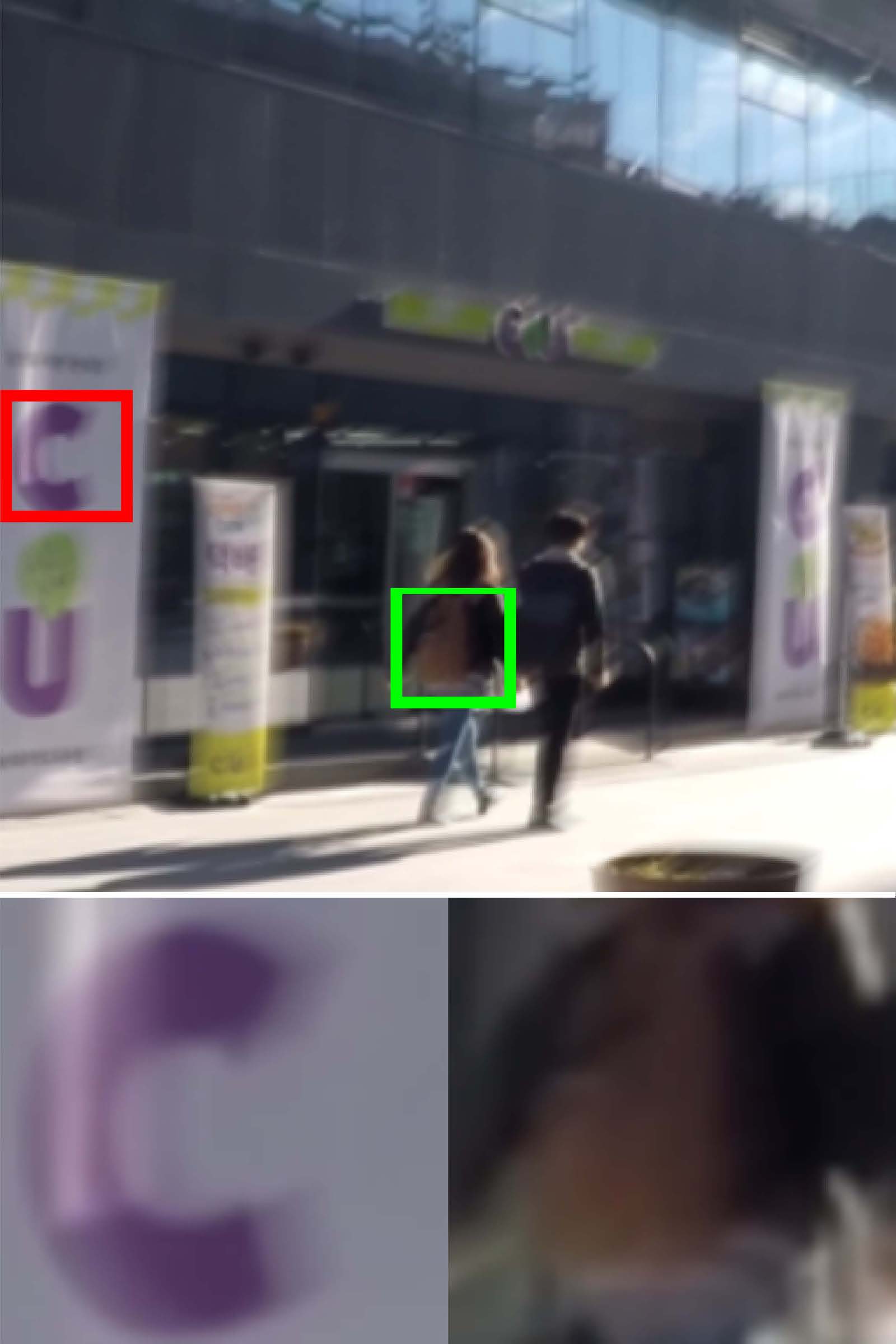}}
\subfigure[Output]{\includegraphics[width=0.32\linewidth]{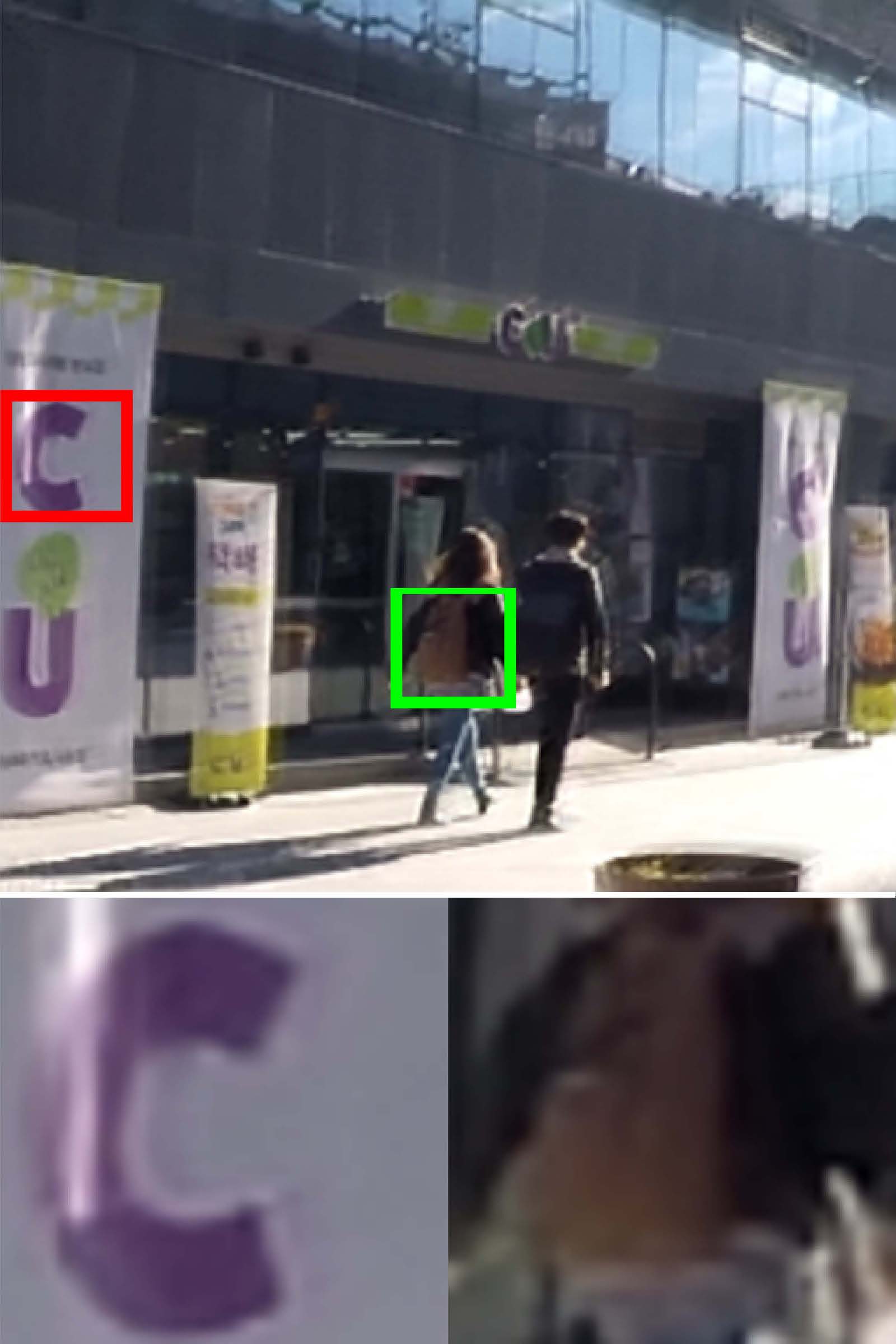}}
\subfigure[Ground truth]{\includegraphics[width=0.32\linewidth]{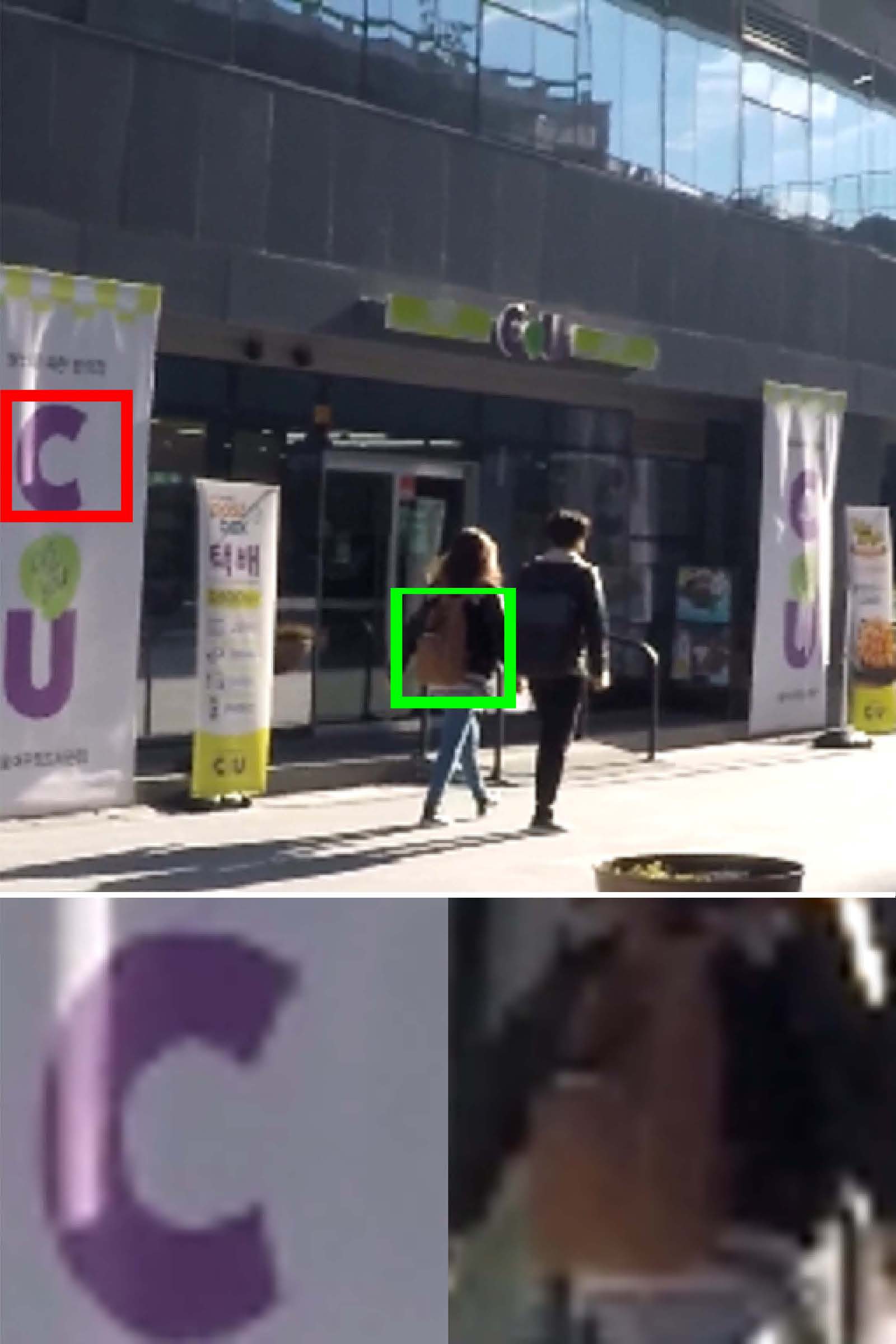}} \\
\caption{Image restoration with our Golf Optimizer on GoPro dataset \cite{nah2017deep}. One can see that the edge of objects are well recovered with rich details.}\label{fig:demo}
\end{center}
\vskip -0.2in
\end{figure}

In this paper, we test our proposed Golf Optimizer network on the benchmark dataset GoPro \cite{nah2017deep}.
In Figure \ref{fig:demo}, experimental results demonstrate that our deep optimizer achieves appealing performance.

\section{Related Work} \label{sec:rltwrk}
Due to the highly ill-posed nature, the image prior modeling plays important role in blind image deconvolution.
Without explicit assumptions on image prior, CNNs are trained with large amount of image pairs, owing to their ability to represent realistic image priors.
However, it's non-trivial to directly use end-to-end CNNs to perform image deconvolution \cite{xu2014deep}.
With the variable splitting technique, several approaches \cite{zhang2017learning, chang2017one, zhang2017ircnn,bigdeli2017deep} train deep CNNs as image prior or denoiser in a plug-and-play strategy.
In these methods, pretrained deep CNNs are integrated as the proximal projector into model-based optimization.
Particularly, in \cite{bigdeli2017deep}, instead of learning image priors, the propagation of prior is learned by a denoising autoencoder.
However, these priors are usually learned independently from the task of deconvolution.

\begin{figure*}[t!]
\vskip 0.2in
\begin{center}
\centerline{\includegraphics[width=0.7\linewidth]{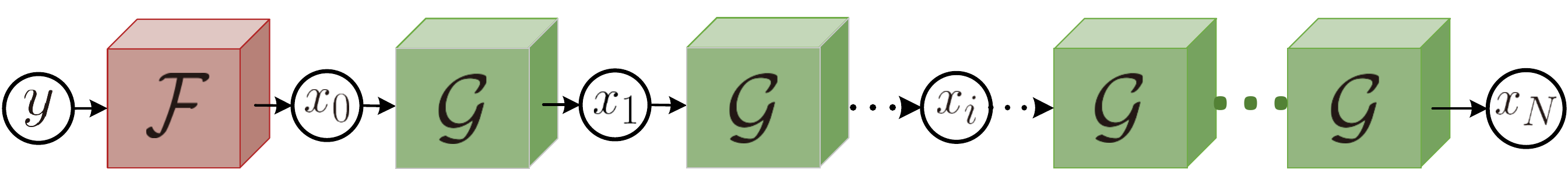}}
\caption{Illustration of the flowchart of \textit{Golf Optimizer}. $\F$ denotes the first network that aggressively propagates from degraded image $y$ towards sharp image, and generates the estimate $x_0$. The sequential optimizers $\G$ which share same parameter of architecture and is referred to the second network, make delicate correction on restoration $x_i$, and generate the final estimate $x_N$ after $N$ iterations.}
  \label{fig:description}
\end{center}
\vskip -0.2in
\end{figure*}

Some works \cite{li2018learning,asim2018deep} learn deep image priors based on generative models, such as generative adversarial networks (GAN) \cite{goodfellow2014generative} and variational auto-encoders (VAE) \cite{kingma2013auto} which are successful in modelling the complex data distribution of realistic images.
In \cite{ramakrishnan2017deep}, a deep densely connected generative network is trained with a Markovian patch discriminator.
DeblurGAN is proposed in \cite{DeblurGAN} where Wasserstein loss is used to circumvent the problems such as model collapse and gradient vanishing.

To handle degenerated images with different levels, a line of works \cite{gong2018learning,liu2019learning,liu2018learning_b} learn CNN optimizers to mimic the propagation of image update in conventional gradient-based optimization.
Particularly, in \cite{gong2018learning}, all main operations of image propagation using gradient descent, including the gradient of image prior, are parameterized with a trainable network for non-blind deconvolution.
With given fixed kernel, the iterative gradient descent update is estimated by the network based only on the last update.
Hence for the recurrent network each estimation remains to be the same task, which limits this architecture to non-blind deconvolution.
In \cite{liu2019learning,liu2018learning_b}, the propagations of image updates are performed by pretrained CNNs which is irrelevant to deconvolution task, and the prior is corrected by optimization-based projection.

\section{Proposed Method} \label{sec:propme}
As mentioned at previous sections, our Golf Optimizer is separated into two sub-modules, in which the first network denoted as $\F$ makes an aggressive propagation towards optimum, and the second network referred to as the deep-optimizer $\G$ employs recurrent structure to iteratively correct our estimation.
Given blurry input $y$, the network $\F$ and $\G$ are performed as $x_0 = \F(y), ~~~  x_i = \G(x_{i-1}), ~~~i=1,...,N$.
%
Figure \ref{fig:description} provides brief description of our optimizer, and the architectural details will be given latter.

Before proceeding, we first define some notations that appear throughout this paper.
We use $\|\cdot\|$ to denote the Euclidean $L_2$ norm, and let the random variable $\epsilon$ follows zero-mean Gaussian distribution with deviation level $\sigma$, i.e., $\epsilon \sim \N(0,\sigma^2 I)$, where $I$ is the identity matrix.
We let $\tilde{x}$ be the sharp image, and $x_0=\F(y)$ be the the output of the first network w.r.t blurry input $y$, and $x_i = \G(x_{i-1})$ be the output of iterative optimizer $\G$ for $i=1,...,N$ for some $N \geq 1$.

\subsection{Problem Formulation}
We start the image deconvolution problem with the MAP estimate where the posterior distribution of restored image $x$ given the blur degradation $y$, is formed as $p(x|y)\propto p(y|x)p(x)$.
The target of image deconvolution is to minimize the negative of the logarithm of the posterior distribution over image space $\C$, formally:
\begin{equation} \label{eq_probform}
  \begin{split}
    &~~~~\min_{x \in \C} \{ -\log p(x|y) \} \\
    &= \min_{x \in \C} \{ - \log p(y|x) - \log p(x) \} \\
    & = \min_{x \in \C} \{\psi(x) = \text{data}(x) + \text{prior}(x)\},
  \end{split}
\end{equation}
where the negative of the logarithm of the likelihood often refers to the data fidelity term, and $-\log p(x)$ referred as prior is the regularization term of model to constrain the deconvolution solution within $\C$.

\subsection{Instability of Priors Learning}

The prior learning is the central component in many image restoration tasks due to their ill-conditional properties.
Though generative models with discriminative supervision has achieved significant improvement in image deconvolution \cite{li2018learning,asim2018deep,ramakrishnan2017deep,DeblurGAN}, there still are some problems in prior learning.
On one hand, if the discriminator behaves badly, then the generator cannot receive any accurate feedback to model image prior.
While the discriminator is expert, the gradient of loss would tend to zero and the prior learning will become slow even jammed.
This phenomenon is known as \textit{gradient vanishing} in GAN \cite{arjovsky2017towards}.
On the other hand, true solution may be far away from the range of the generators, and the pretrained generators may not precisely model the distribution of realistic images \cite{asim2018deep,bora2017compressed}.

Instead of learning prior by generative models, we employ a residual unit (i.e., identity shortcut) as our deep-optimizer $\G$ to learn the gradient of prior.
Since that in the formulation the residual unit coincides with the gradient descent method, we \textit{asymptotically learn the descent direction of prior as deep-prior via residual learning} from identity shortcut as,
\begin{equation} \label{eq:grad_prior_missing_sigma}
  -\nabla \text{prior}(x)= \G(x) - x,
\end{equation}
which will be stated formally in section \ref{sec:post_net}.

Analogous works \cite{bigdeli2017deep,jin2017noise,bigdeli2017image} build on denoising autoencoders (DAE) to learn the gradient of prior, which is referred to as deep mean-shift prior.
The pretrained DAE is integrated into the optimization as the regularization term in a plug-and-play strategy, which is limited to non-blind deconvolution.
Moreover, regarding to the generalization of prior learning, the distribution of images which is trained to build DAE in denoising may not coincide with that of images in deconvolution.

Unlike these methods, we train our deep-optimizer to capture the prior within the blind deconvolution task, to make sure the optimizer learn deconvolution related priors.
Formally, for each pair of $(x,\tilde{x})$ we train our optimizer $\G$ by minimizing
\begin{equation} \label{eq:ojb_train_G}
  \int_{\tilde{x}} { \frac{1}{N} \sum_{i=1}^{N} \left \|\G^{(i)}(x) - \tilde{x} \right\|^2 p(\tilde{x}) } \mathrm{d} \tilde{x},
\end{equation}
where $x$ is the degraded image and $N$ is the total iteration of optimizer $\G$.
In Eq.\ref{eq:ojb_train_G}, $\G^{(i)}(\cdot)=\G \circ \cdots \circ \G(\cdot)$ denotes the $i$-fold composition of $\G(\cdot)$ where $\circ$ denotes the composition operator.

\begin{figure}[t!]
\vskip 0.2in
\begin{center}
\subfigure[Discriminative prior]{\includegraphics[width=0.48\linewidth]{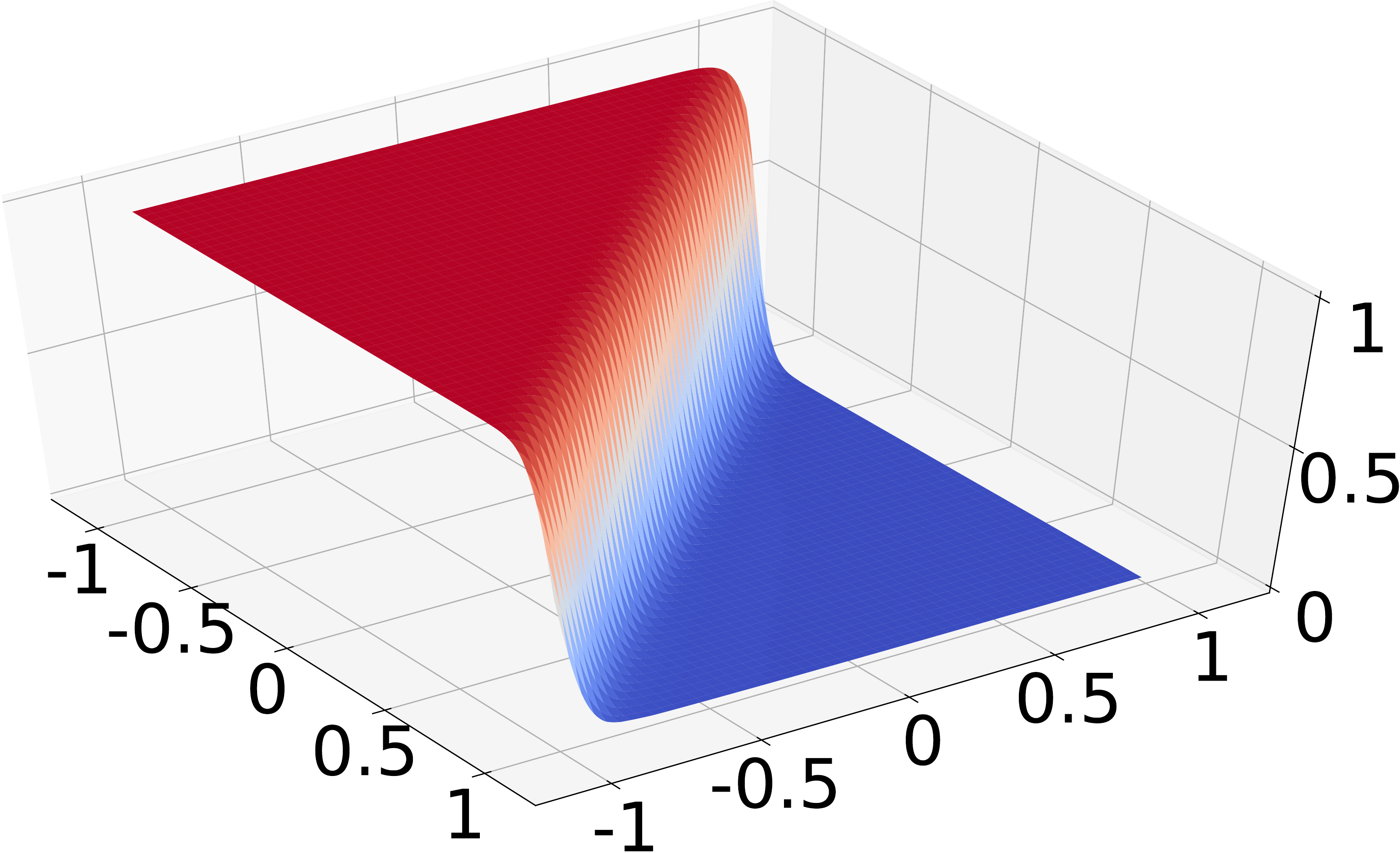}
\label{fig:dis_prior}
}
\subfigure[Gradient constrained prior]{\includegraphics[width=0.48\linewidth]{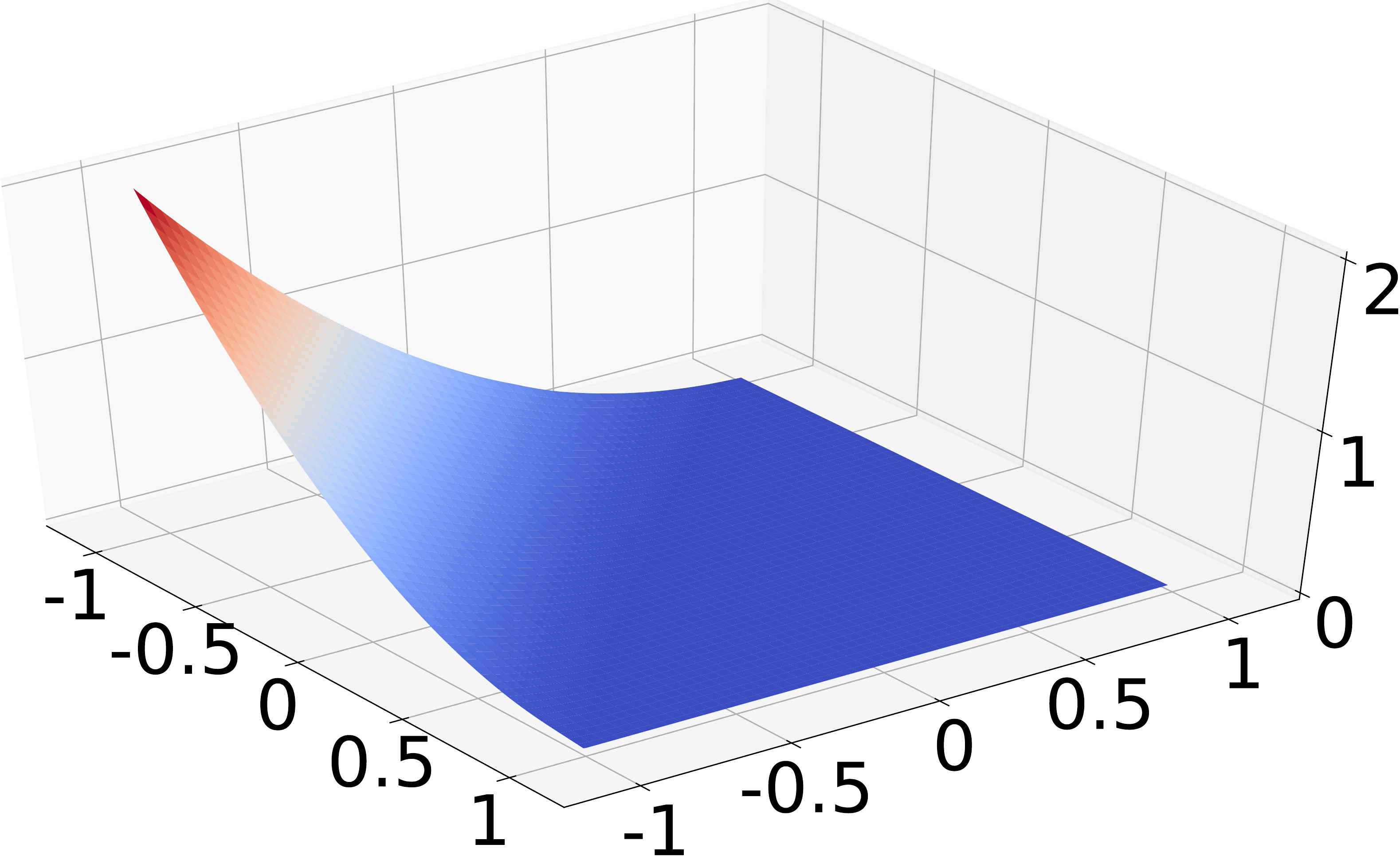}
\label{fig:our_prior}
} \\
\caption{Advantage of gradient learning for prior. Suppose we are to restore image $u$ consisting of only $2$ pixels: $u=(u_1,u_2)$. An image is said to be sharp if $u_1+u_2>0$, otherwise blur. The ground truth $\tilde{u}$ of a blur image $u$ is defined as the symmetric point w.r.t line $u_1+u_ 2=0$. (a) We trained a MLP classifier $p$ to distinguish blurry/sharp images so that $p(\tilde{u})=\textcolor{blue}{\bf 0}$ and $p(u)=\textcolor{red}{\bf 1}$. When using $p$ as prior to deblur $u$, we can see the gradient are mostly zero. (b) we directly train a network $f(u)$ to learn the gradient of the optimal prior $\nabla p^* =u-\tilde{u}$, where the $p^*$ is obtained from $f$ via integration $p^*=\int f$. We can see the gradients are non-zero and point toward the ground-truths.
}
\label{fig:GradPriorExp}
\end{center}
\vskip -0.2in
\end{figure}

The most significant superiority of learning the gradient of prior is that it can eliminate the instability of training with discriminative criterion.
The instability lies on the improper evaluation metric $\L$ on the performance of restoration, which propagates unstable gradient $\frac{\partial \L}{\partial x}$ of prior back to generator.
Figure \ref{fig:dis_prior} gives an simple example to demonstrate gradient vanishing with pretrained prior.
To address this instability, we train $\G$ with Eq.\ref{eq:grad_prior_missing_sigma} to learn the gradient of prior by minimizing Eq.\ref{eq:ojb_train_G}, which reveals that learning prior is equivalent to the minimization of
\begin{equation} \nonumber
  \int_{\tilde{x}} \left\| \nabla \text{prior}(x) - (x - \tilde{x}) \right\|^2 p(\tilde{x})\mathrm{d} \tilde{x}.
\end{equation}
This shows the gradient of prior learns the difference between the degeneration and sharp image, which essentially eliminates the instability of learning prior gradient.
Figure \ref{fig:our_prior} provides the advantage of learning gradient of priors.

\subsection{Catastrophic Forgetting in Iterative Blind Deconvolution}
Another important issue in blind deconvolution with recurrent network involves the catastrophic forgetting problem when trivially applying optimizer iteratively for image restoration.
Considering the optimization with gradient descent as minimization of the objective $\psi(x)$ in Eq.\ref{eq_probform}, the restored solution is iteratively updated as
\begin{gather}
  x_{\text{mid}}  = x_i - \alpha \nabla \text{data}(x_i), \nonumber \\
  x_{i+1}  = \arg \min_{x \in \mathcal{C}} \left\{ \frac{1}{2\beta} \|x - x_\text{mid} \|^2 + \text{prior}(x) \right\} \label{eq:prox},
\end{gather}
where Eq.\ref{eq:prox} typically refers to proximal operation of $\text{prior}(x)$ with controlling factor $\beta$.
And it is identical to $x - \beta \nabla \text{prior}(x)$ when $\text{prior}(x)$ is differential to $x$.
This update form includes two modules on image $x$, gradient descent module and prior projection module respectively.

\begin{figure}[t!]
\vskip 0.2in
\begin{center}
\subfigure[\textit{Task1 after Task2}]{\includegraphics[width=0.49\linewidth] {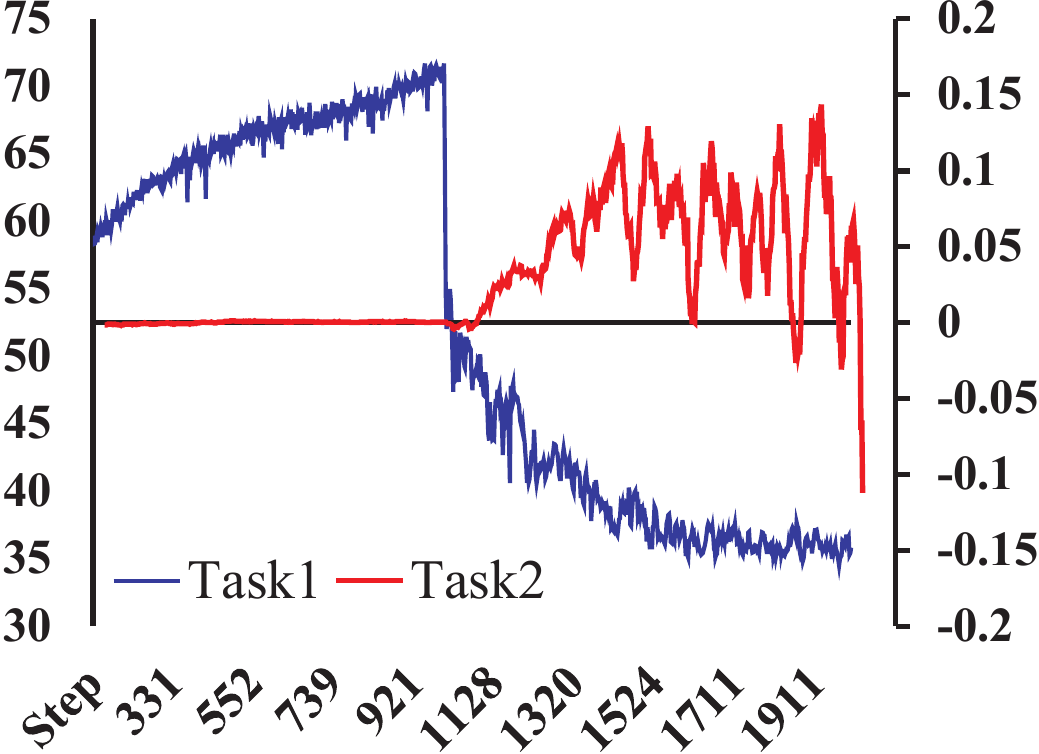}}
\subfigure[\textit{Task2 alone}] {\includegraphics[width=0.48\linewidth] {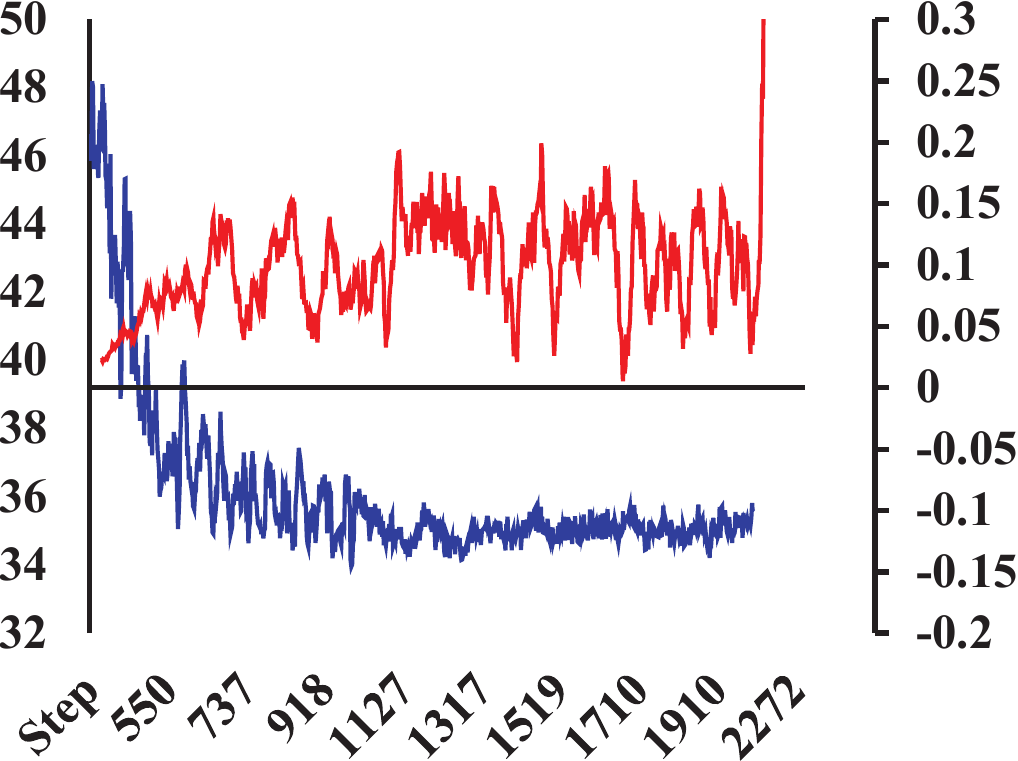}} \\
\subfigure[\textit{Task2 after Task1}]{\includegraphics[width=0.49\linewidth] {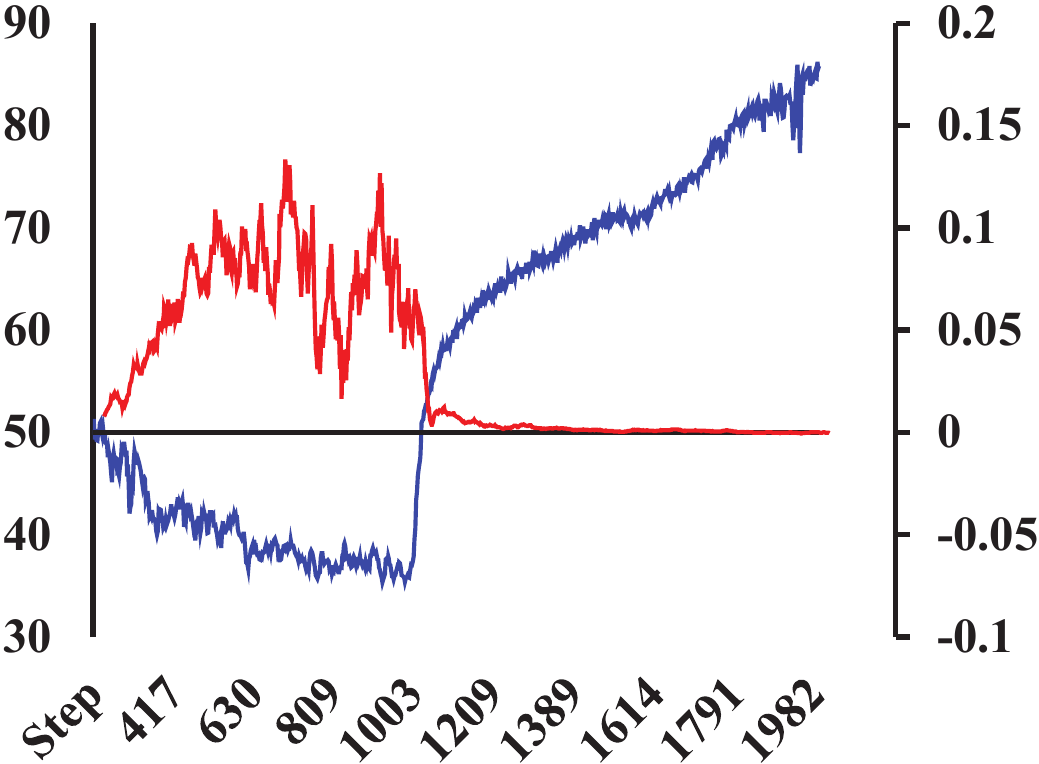}}
\subfigure[\textit{Task1\&2 simultaneously}]{\includegraphics[width=0.47\linewidth] {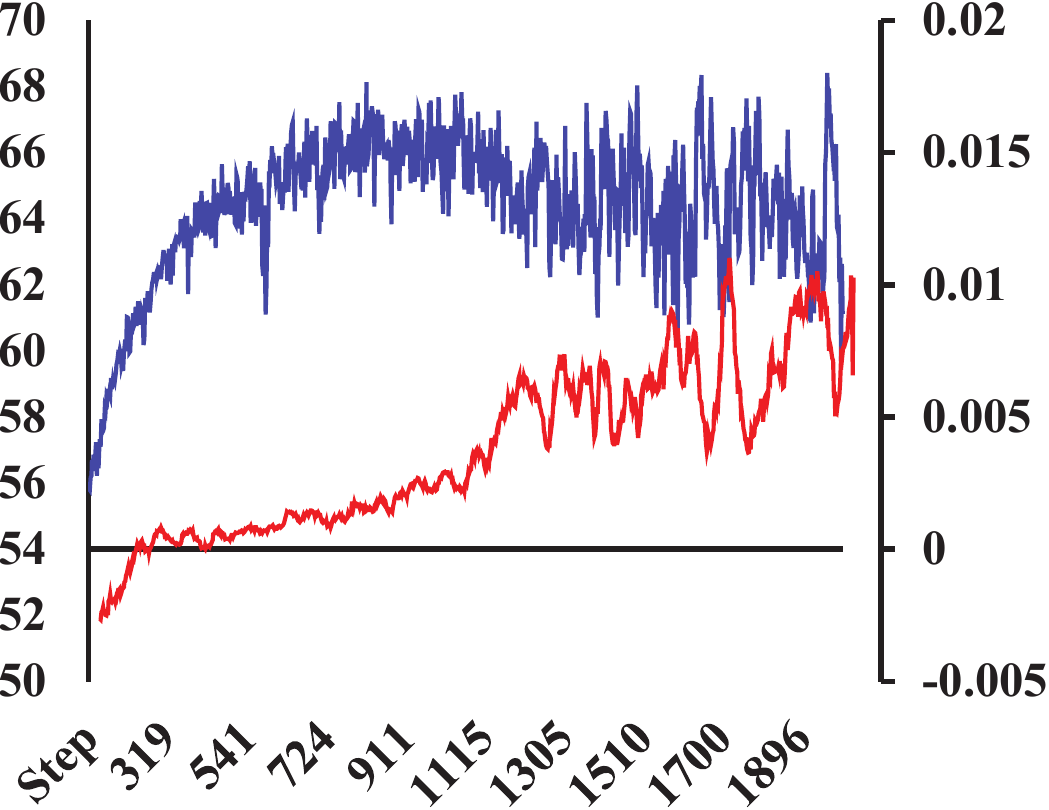}}
\caption{Catastrophic forgetting in blind deconvolution in training optimizer $\F$. Task1 stands for training on sharp image $\tilde{x}$ with PSNR of $\F(\tilde{x})$ in the left Y-axis, while Task2 represents for training on blurry image with the PSNR of $\F(x)$ minus that of $y$ in the right Y-axis. In (a) and (c), training on the second task after on the first significantly decrease the performance of optimizer on the first task. (b) training Task2 alone will mildly promote restoration. (d) simultaneous training on both task with more computation hardly improves the deconvolution.}
\label{fig:forget}
\end{center}
\vskip -0.2in
\end{figure}

For non-blind image deconvolution with end-to-end network, RGDN \cite{gong2018learning} integrates these two modules into a recurrent convolutional network, where the gradient of image prior unit is replaced by a common CNN block.
Given fixed kernel, the iterative update of restoration is performed as $x_{i+1} = \F(x_i;\theta)$ with network $\F$ parameterized by $\theta$.
And the update on $x_{i+1}$ with $\F$ remains the same task.
However, this particular recurrent architecture is limited to non-blind deconvolution, in which the updates share the same $\F$.

For blind image deconvolution, considered at single update on blurry image $x_i$ as $x_{i+1} = \F(x_i;\theta)$ with some unknown kernel $k$, the restored image $x_{i+1}$ would follow some other unknown degeneration which may even not be accordant with physical process of degradation.
In this situation, those intermediate data $\{x_i|x_i = \F(x_{i-1})\}$ estimated by network $\F$ would change the distribution of image degeneration, and this would lead network $\F$ to multitasking, which would cause \textit{catastrophic forgetting} \cite{Kirkpatrick3521,serra2018overcoming,lee2017overcoming}.
It is not pleasing to train deconvolution network $\F$ to accommodate new knowledge, which makes the training of network more tough.
Figure \ref{fig:forget} demonstrates the difficulties of training network for blind image deconvolution.

To address the issue of catastrophic forgetting, we simply adopt an vanilla residual CNN $\F$ to make an aggressive propagation towards the optimum, and fed the output to $\G$ for further delicate adjustment.
Intuitively we need to ensure that, for $i,j \geq 0$ and $i \neq j$, the distribution of $x_i$ and that of $x_j$ should follow same distribution by slightly different distribution parameters.
According to Eq.\ref{eq:ojb_train_G}, we have for each $x_{i}$, $x_{i+1} = \G(x_i)$ follow identical distribution with $x_i$.
Hence, we only need to ensure $x_0$ follows same distribution with $\{x_i| i>0\}$.
To achieve this, we train $\F$ using a similar target to $\G$:
\begin{equation} \label{ojb_train_F}
  \int_{\tilde{x}} \left\| \F(y) - \tilde{x}  \right\|^2 p(\tilde{x}) \mathrm{d}\tilde{x}.
\end{equation}

\section{Learning Deep Priors via Optimizer}\label{sec:post_net}
The aim of our deep-optimizer is to utilize deep-priors to make adjustment to the restored images.
To achieve this, we adopt residual structure for the proposed optimizer.
In this section, we will show that the deep-optimizer is capable of learning the gradient of prior terms.
Also, we will elaborate our choice of residual architectures via formal deductions.

\subsection{Learning Gradient of Prior}

We first denote $g_\sigma(\cdot)$ as the zero-mean Gaussian distribution with deviation level $\sigma$.
Our formulation is low-bounded by the logarithm of the MAP estimator \cite{jin2017noise,bigdeli2017deep} as
\begin{equation} \nonumber
  \begin{split}
    &~~~~ \max_{x} \log p(y|x) p(x) \\
    &\geq \max_x \left\{ \log p(y|x) + \log \int p(x + \epsilon) g_{\sigma}(\epsilon) \mathrm{d}\epsilon \right\},\\
  \end{split}
\end{equation}
where the prior is expressed as the logarithm of the Gaussian-smoothed distribution $p(x)$ as:
\begin{equation} \label{eq:prior}
  -\text{prior}(x) = \log \int p(x + \epsilon) g_{\sigma}(\epsilon) \mathrm{d}\epsilon.
\end{equation}

As in many low-level image enhancements e.g., JPEG deblocking, super-resolution, denoising, we model image degradation with the difference $\epsilon$ between the ground-truth $\tilde{x}$ and the observed degradation $x$ \cite{zhang2017beyond,nah2017deep}, i.e., $x = \tilde{x} - \epsilon$.
For the sake of analytical convenience, we further assume that the difference $\epsilon$ follows Gaussian distribution with unknown deviation level $\sigma$ as $\epsilon \sim \N(0,\sigma^2 I)$.
Ideally, we rewrite Eq. \ref{eq:ojb_train_G} to minimize the following objective
\begin{equation} \label{eq:inf_obj_train}
  \int_{\tilde{x}} \lim_{N \rightarrow \infty} \frac{1}{N} \sum_{i=1}^{N} \left \|\G^{(i)}(x) - \tilde{x} \right\|^2 p(\tilde{x}) \mathrm{d}\tilde{x}.
\end{equation}
Replacing the operation of averaging on composition by taking expectation w.r.t random variable $\epsilon$ and using $\tilde{x} = x+\epsilon$, we can rewrite Eq.\ref{eq:inf_obj_train} as
\begin{equation} \nonumber
  \int_x \E_{\epsilon \sim \N(0,\sigma^2 I) } \left[  \left \|\G(x) - (x+\epsilon) \right\|^2 p(x+\epsilon) \right] \mathrm{d}x,
\end{equation}
which can be differentiated w.r.t $\G$ and set to be equal to $0$.
By denoting the optimum as $\G^*(x)$, we obtain
\begin{equation} \nonumber
  0 = \E_{\epsilon \sim \N(0,\sigma^2 I) } \bigg[ \big( \G^*(x) - (x+\epsilon) \big) p(x+\epsilon) \bigg].
\end{equation}
This leads to our final optimizer $\G^*(x)$ as
\begin{equation} \nonumber
  \G^*(x) = \frac{\E_{\epsilon \sim \N(0,\sigma^2 I) }[p(x+\epsilon)(x+\epsilon)]}{\E_{\epsilon \sim \N(0,\sigma^2 I)} [p(x+\epsilon)] }.
\end{equation}
Following \cite{bigdeli2017deep,alain2014regularized}, the gradient of prior can be learned by $\G$ with
\begin{equation} \label{eq:grad_prior}
  \begin{split}
    -\nabla \text{prior}(x) &= \nabla \log \int p(x+\epsilon) g_\sigma(\epsilon) \mathrm{d}\epsilon \\
    &= \frac{1}{\sigma^2}\left( \G^*(x) - x \right).
  \end{split}
\end{equation}

\textbf{Missing $\sigma^2$}:
When the value of $\sigma$ is unknown, we can modify Eq.\ref{eq:grad_prior} by dropping $\sigma^2$.
For a trained optimizer $\G$, we just take the quantity $\G^*(x) -x$ and the gradient of prior should approximately be the score up to a multiplicative constant \cite{alain2014regularized}.
Together with the assumption on $\epsilon$, the gradient of prior can be asymptotically learned from $\G$, which forms our core ingredient in deep-prior learning for blind image deconvolution.

\subsection{The Choice on Architecture of Optimizer}
The deep-optimizer $x_i = \G(x_{i-1})$ aims at achieving descent objective of the difference between current estimate and optimal, formally
\begin{equation} \label{eq:goal}
  \|\G(x_i) - \tilde{x}\|^2 \leq \|\G(x_{i-1}) - \tilde{x}\|^2 + \eta,
\end{equation}
where $\eta \geq 0$ is a small non-negative constant.
We denote the objective as $\L(\G,x_i,\tilde{x})$, and take a second-order Taylor expansion on $\L(\G,x_i,\tilde{x})$ around $x_{i-1}$:
\begin{equation} \nonumber \small
  \begin{split}
    &~~~~~~\L(\G,x_i,\tilde{x}) = \|\G(x_i) - \tilde{x}\|^2 \\
    &=\|\G(x_{i-1}) - \tilde{x}\|^2 + 2\left\langle (\G(x_{i-1})-\tilde{x})^T \nabla \G(x_{i-1}), x_{i}-x_{i-1} \right\rangle \\
    &~~~+(x_{i} - x_{i-1})^T \nabla^2 \G(x_{i-1}) (x_{i} - x_{i-1}) + o \left( \|x_{i} - x_{i-1}\|^2 \right),
  \end{split}
\end{equation}
where $o(\cdot)$ denotes the remainder term.
Combing above with Eq.\ref{eq:goal} and letting $\Delta_i=\G(x_{i-1}) - x_{i-1}$, we obtain:
\begin{equation}\nonumber
  \begin{split}
    \eta &\geq 2\left\langle (\G(x_i)-\tilde{x})^T \nabla \G(x_i), \Delta_i) \right\rangle \\
    &~~~~~~+\Delta_i^T \nabla^2 \G(x_i) \Delta_i + o \left( \|\Delta_i\|^2 \right).
  \end{split}
\end{equation}
This shows the term $\|\Delta_i\|$ tends to be $o(\|\eta\|)$, which motivates us to pick up residual learning \cite{he2016deep}.
Unlike the deep residual network \cite{he2016deep} consisting of residual units in which identity shortcuts is used only inside the units, the iterative network $\G$ employs a single residual unit to learn the residual mapping from current estimate to next one, i.e. $\G(x) = x + r(x)$, where $r(\cdot)$ is the residual mapping of $\G$.
With missing term $\sigma^2$ in Eq. \ref{eq:grad_prior}, the optimization with gradient descent as minimization the objective $\psi(x)$ is
\begin{equation}
  \begin{split}
    x_{i+1} &= x_i - \nabla \text{prior}(x_i) = \G(x_i)= x_i + r(x_i).
  \end{split}
\end{equation}
Hence, the gradient of prior can be learned from the residual mapping by applying recurrent optimizer.

\section{Implementations}
\begin{figure}[t!]
\vskip 0.2in
\begin{center}
\centerline{\includegraphics[width=0.7\linewidth]{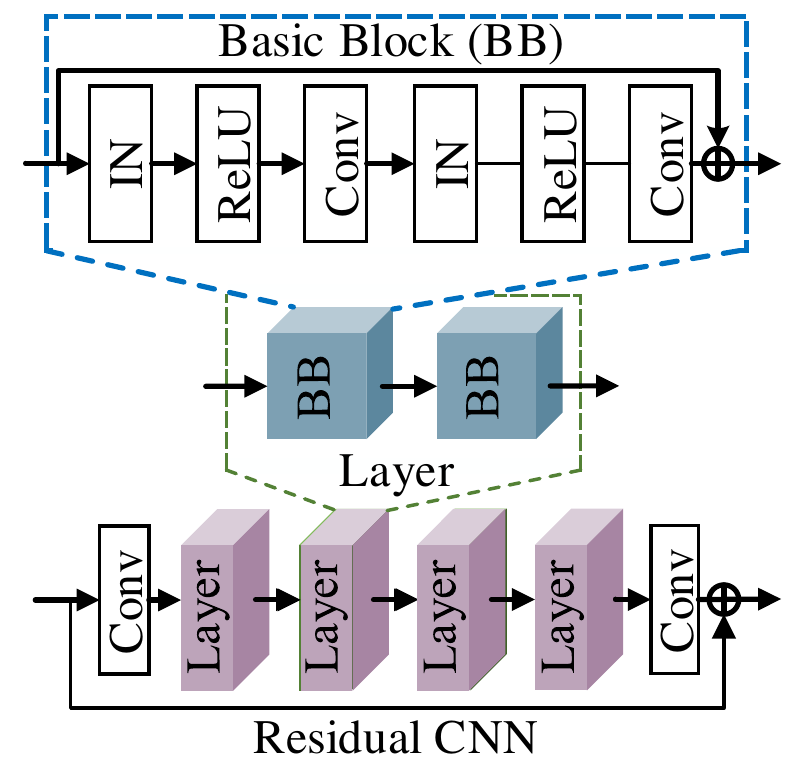}}
\caption{The residual CNN architecture. The kernel size of the first/last layers are fixed to be 7 and 3. Inside BB unit, the kernel size of the Conv layers are 7 and 5. The filters in the network is a tunable hyper-parameter. IN is Instance-Normalization layer.}
\label{fig:arch}
\end{center}
\vskip -0.2in
\end{figure}

\subsection{Architecture}
Here we give our network architectures of Golf Optimizer.
$\F$ and $\G$ have identical network architecture but different configurations. The architecture is shown in Figure \ref{fig:arch}.
In the network, there are four Layers, each consisting of two Basic-Block (BB) units.
Inside BB is the residual structure, where residual mapping is stacked by six basic layers shown in white rectangles.
Except for the first/last layers whose input/output channels are 3, all the other convolution layers have identical in/out channels, thus having same number of filters $\mathfrak{f}$.
We set $\mathfrak{f}=32$ for $\F$, and $\mathfrak{f}=16$ for $\G$. All the Conv layers have stride $=1$, making feature-maps the same spatial size as input. Table \ref{tab:net_gf} provides the details of the network.



\begin{table}[h!]
\caption{Implementation of Network $\F$ and $\G$. Layer type is followed by the filter settings of contained Conv layers formatted as (in-channels $\rightarrow$ out-channels). Bottom row lists the total number of basic layers (i.e. Conv, ReLU, etc.) }
\label{tab:net_gf}
\vskip 0.15in
\begin{center}
\begin{small}
\begin{sc}
\begin{tabular}{cc|cc}
\toprule
\multicolumn{2}{c|}{$\G$}  &\multicolumn{2}{c}{$\F$} \\
\midrule
 Layer type & params & Layer type & params           \\
\midrule
          $\textsc{Conv}_{(3\rightarrow 16)}$                & 2,352   & $\textsc{Conv}_{(3\rightarrow 32)}$                & 4,704   \\
          $\textsc{BB}_{(16\rightarrow 16)}   \times 2$      & 38,016  & $\textsc{BB}_{(32\rightarrow 32)}   \times 2$      & 151,808 \\
          $\textsc{BB}_{(16\rightarrow 16)}   \times 2$      & 38,016  & $\textsc{BB}_{(32\rightarrow 32)}   \times 2$      & 151,808 \\
          $\textsc{BB}_{(16\rightarrow 16)}   \times 2$      & 38,016  & $\textsc{BB}_{(32\rightarrow 32)}   \times 2$      & 151,808 \\
          $\textsc{BB}_{(16\rightarrow 3)}   \times 2$       & 3,347   & $\textsc{BB}_{(32\rightarrow 3)}   \times 2$       & 5,779   \\
          $\textsc{Conv}_{(3\rightarrow 3)}$                 & 441     & $\textsc{Conv}_{(3\rightarrow 3)}$                 & 441     \\
\midrule
50 layers              & 120,188 &        50 layers                 & 466,348 \\
\bottomrule
\end{tabular}
\end{sc}
\end{small}
\end{center}
\vskip -0.1in
\end{table}

\subsection{Training}
Our model is trained on the GoPro dataset \cite{nah2017deep}.
There are 3214 pairs of images in the dataset.
Each pair contains a sharp ground truth image and a motion-blurred image.
The size of the raw images in the dataset is $720 \times 1280 \times 3$.
Among the 3214 pairs, 2103 pairs are selected for training purpose.
The training set is generated by randomly cropping $256\times 256$ patches from those images.
Routinely, we applied data augmentation procedures to the training set.
The procedures are geometric transformations, including randomized vertical flipping and horizontal flipping.

As for the parameter optimization of the model, we adopted two loss metrics: content loss and perceptual loss.

\textbf{Content loss}:
MSE loss is widely applied in optimization objectives for image restoration.
Using MSE, content loss function in our objective is defined as
\begin{equation} \nonumber
    \L_c = \frac{1}{M} \| x - \tilde{x} \|^2,
\end{equation}
where the pixel-wise errors between estimated image $x$ and ground truth image $\tilde{x}$ is computed, divided by the number of pixels $M$.

\textbf{Perceptual loss}:
It has been shown that using MSE content loss as sole objective would lead to blurry artifacts due to the pixel-wise average of possible solutions in the solution space \cite{ledig2017photo}, and could potentially cause the distortion or loss of details \cite{yang2018low}.
To alleviate this problem, we utilize another loss metric \emph{perceptual loss} \cite{johnson2016perceptual}. 
Given a trained neural network $\P$, the perceptual loss between $x$ and $\tilde{x}$ is defined as
\begin{equation} \nonumber
   \L_p = \frac{1}{whd}\| \P(x) - \P(\tilde{x}) \|^2,
\end{equation}
where $w,h$ and $d$ are the width, height and depth of the feature map.
In our implementation, $\P$ is set to VGG \cite{simonyan2014very} network.
The output $\P(\cdot)$ is defined to be the 5-th maxpooled feature map of VGG-11 model with batch normalization.


\textbf{Overall Loss Function}:
In the training process, content loss and perceptual loss are combined to form an overall loss function as
\begin{equation} \nonumber
  \L = \L_c + \lambda \L_p ,
\end{equation}
where $\lambda$ is the hyper-parameter controlling the balance between two loss terms.
The same overall loss function is used in the training of both network $\F$ and $\G$.


\begin{figure}[t!]
\vskip 0.2in
\begin{center}

 \quad  Blur input  \quad \quad \quad \quad Output  \quad \quad \quad   Ground Truth \\

\includegraphics[width=0.32\linewidth]{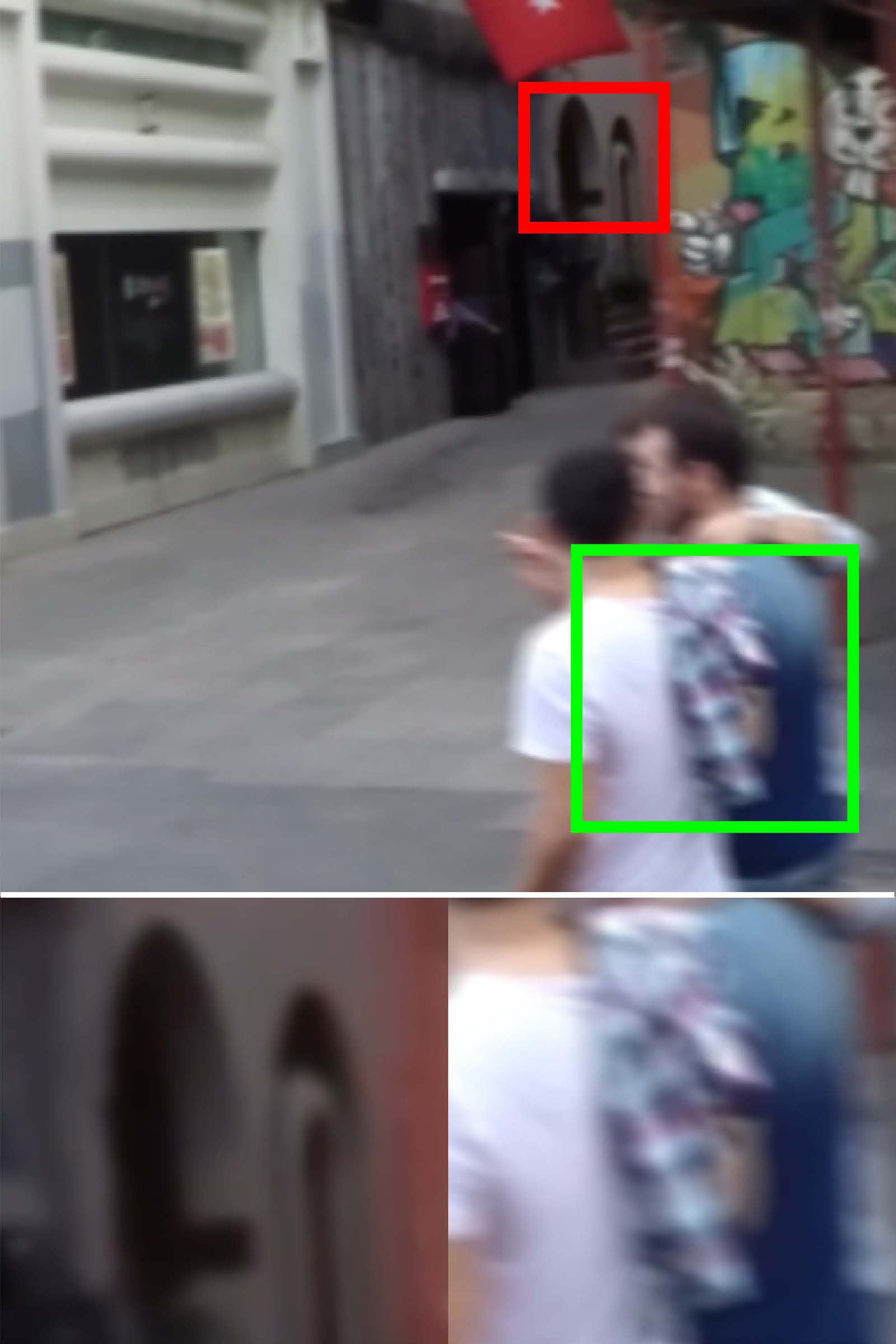}
\includegraphics[width=0.32\linewidth]{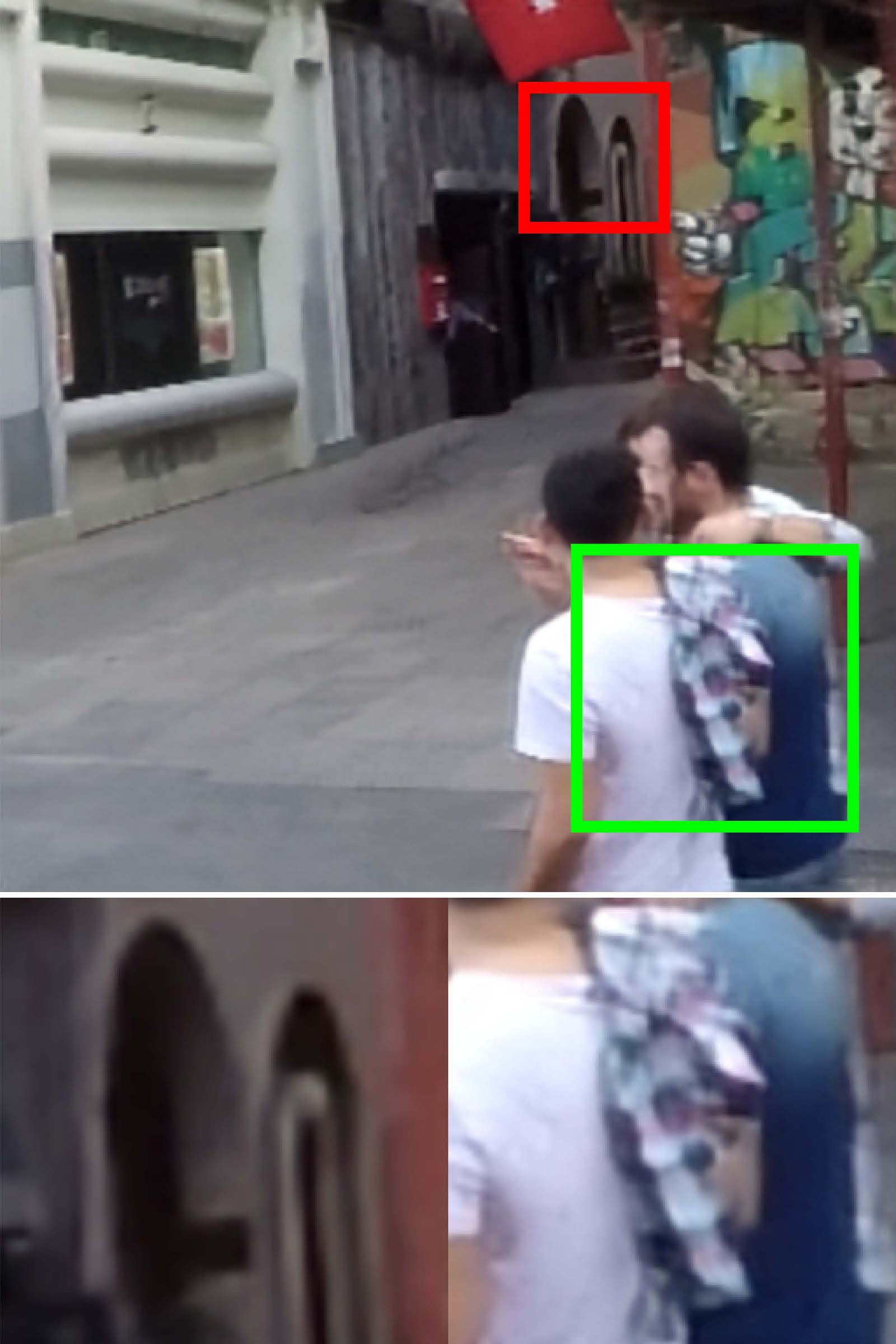}
\includegraphics[width=0.32\linewidth]{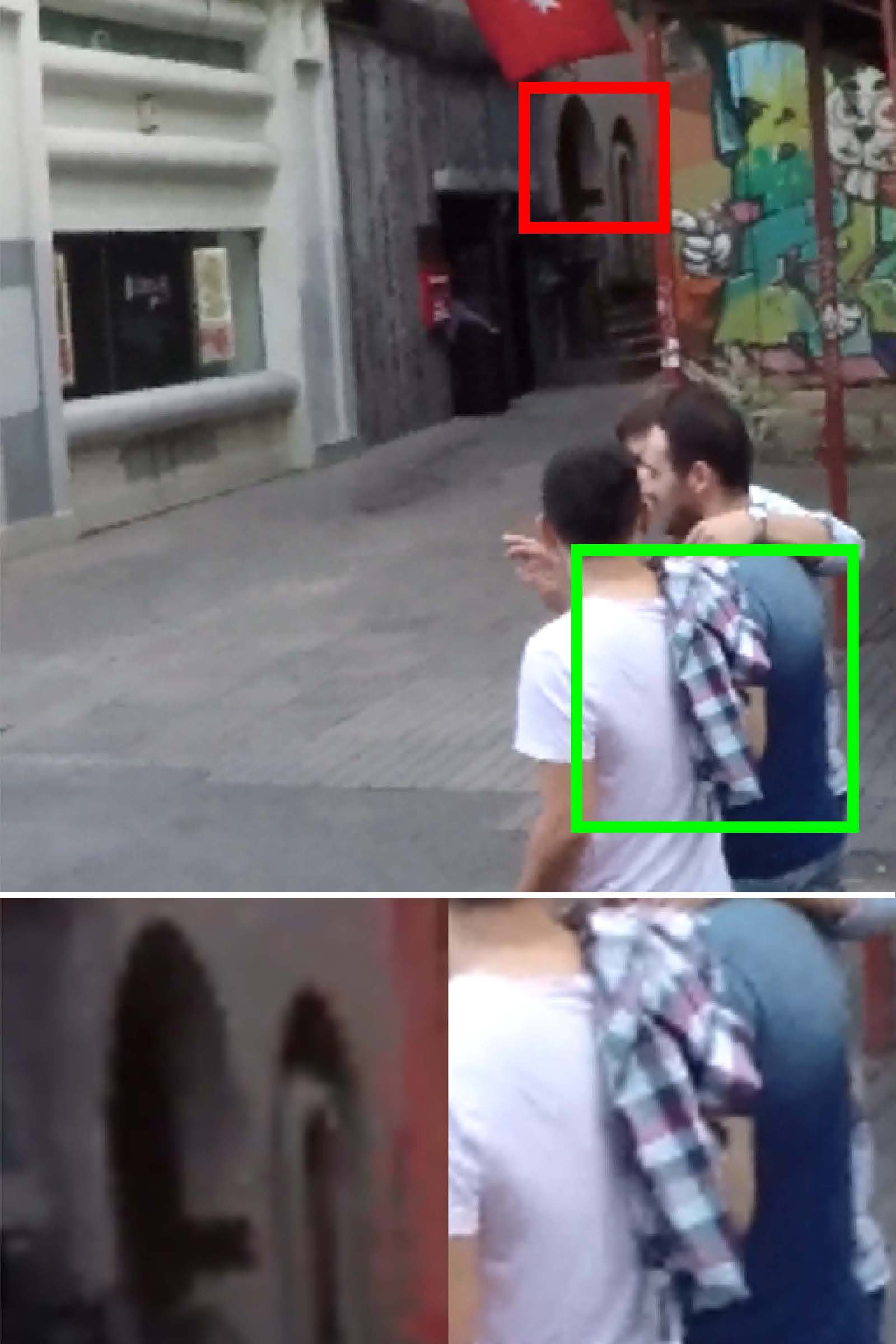} \\
\vspace{1em}
\includegraphics[width=0.32\linewidth]{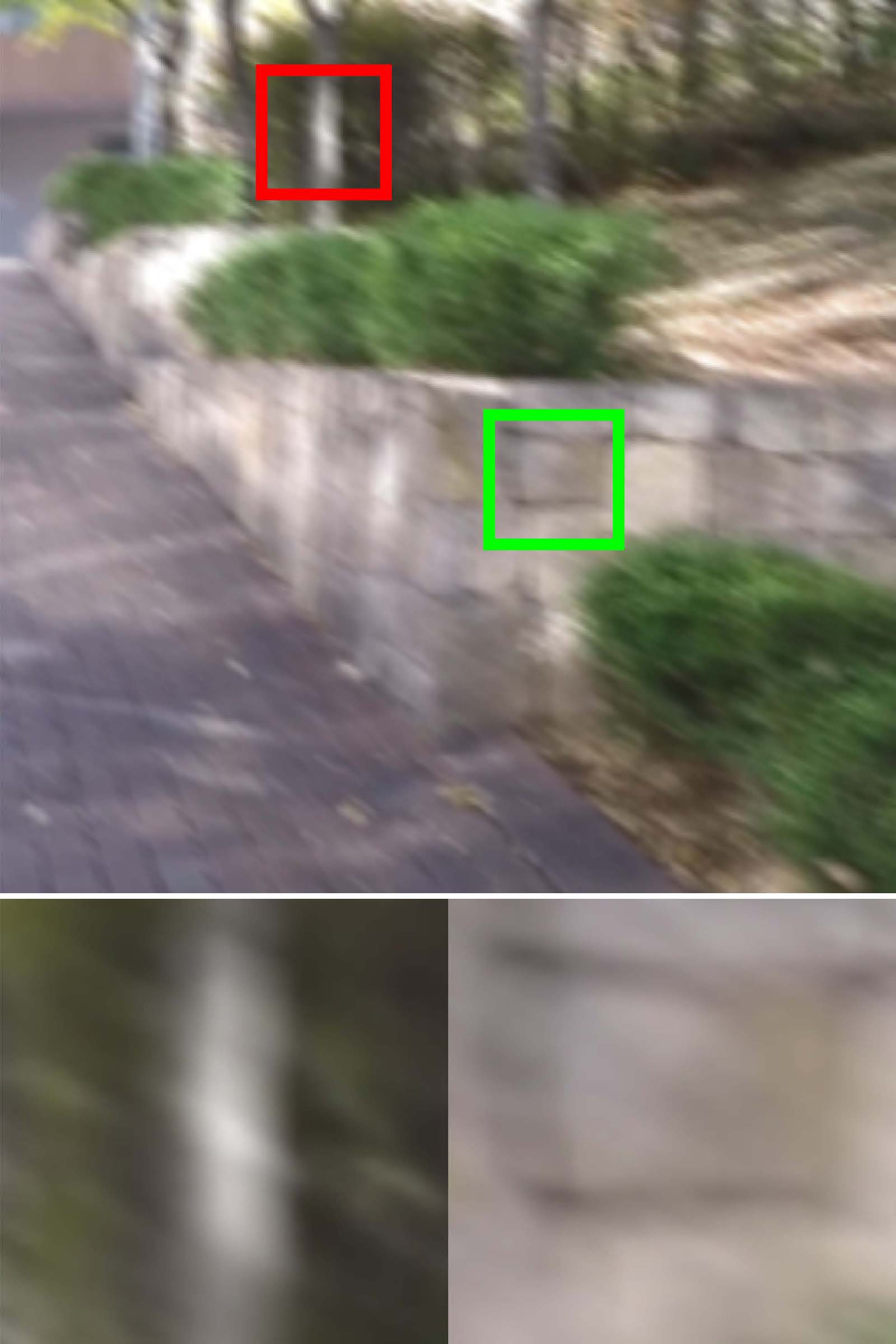}
\includegraphics[width=0.32\linewidth]{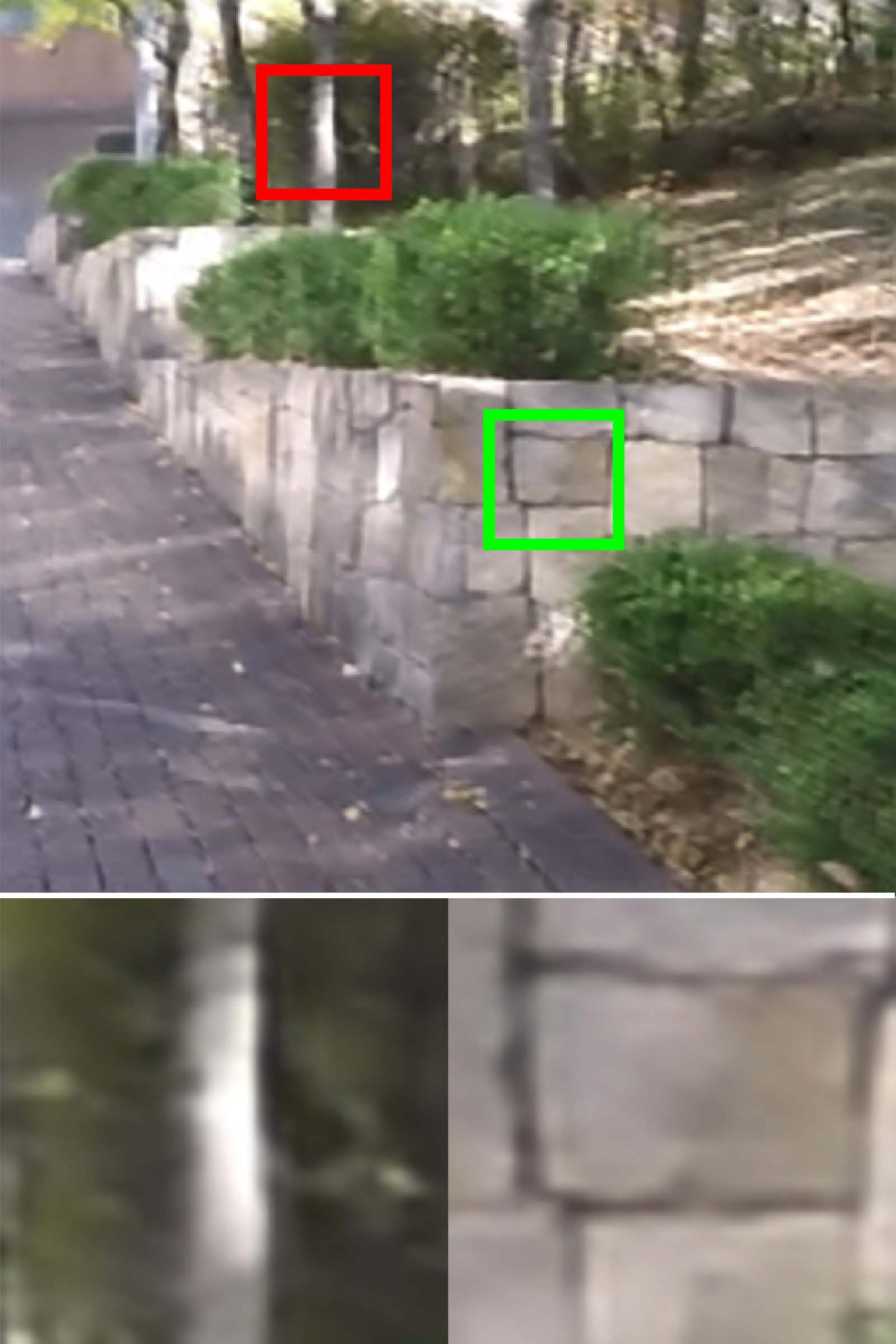}
\includegraphics[width=0.32\linewidth]{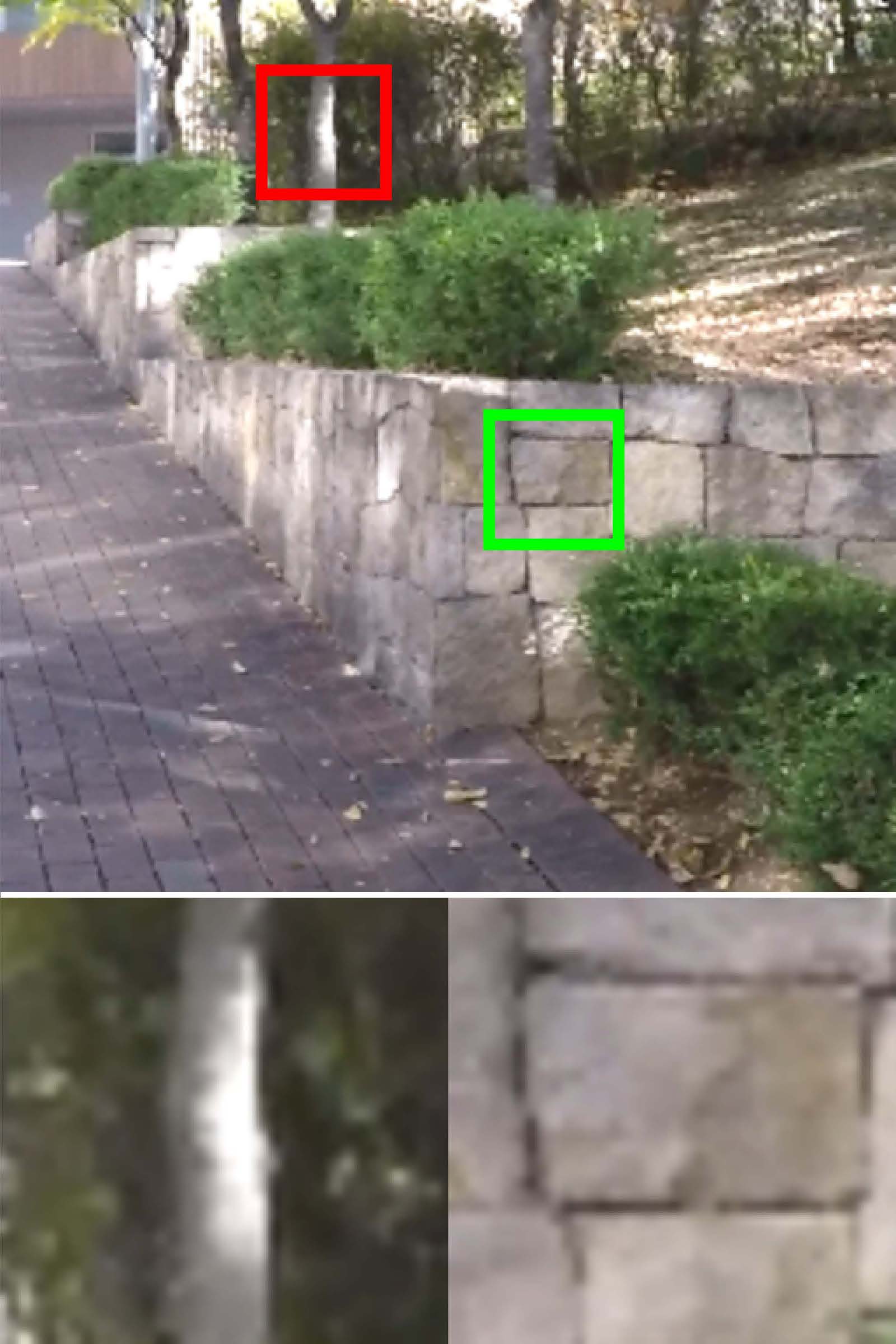}

\caption{Example Results. The PSNR and SSIM values of the output image in the 1st/2nd row is 36.08,26.91, and the corresponding SSIM values are 0.93,0.73.}\label{fig:examples}
\end{center}
\vskip -0.2in
\end{figure}

\section{Experimental Results}
We evaluate the performance of our model quantitatively on GoPro dataset with visual examples. We also provide details of our experiment with analysis.

\subsection{Details of The Experiment}
We train the networks using GTX 1080 Ti graphic card with 11-GB memory. The computer has two Xeon Gold 5118 CPUs and 64 GB RAM.

\textbf{\small Optimization Setting}:
ADAM \cite{kingma2015adam} optimizer is used with $\beta_1=0.9,\beta_2=0.999$ and a mini-batch of size 4. The learning rate is $10^{-3}$ and will exponentially decay to $\gamma^{-11}$ with $\gamma=0.3$.
The $\F$ network is trained for $500,000$ iterations.
Then with $\F$ fixed, $\G$ network is trained to convergence which takes $200,000$ s.pdf.

Ideally, the iteration depth $N$ is unlimited. However, in practice we found using large $N$ would significantly challenge our hardware capacity.
Finally, we set $N=5$ in the experiment.

\begin{figure}[t!]
\vskip 0.2in
\begin{center}
\subfigure[ $\lambda=1e^{-10}$]{\includegraphics[width=0.3\linewidth]{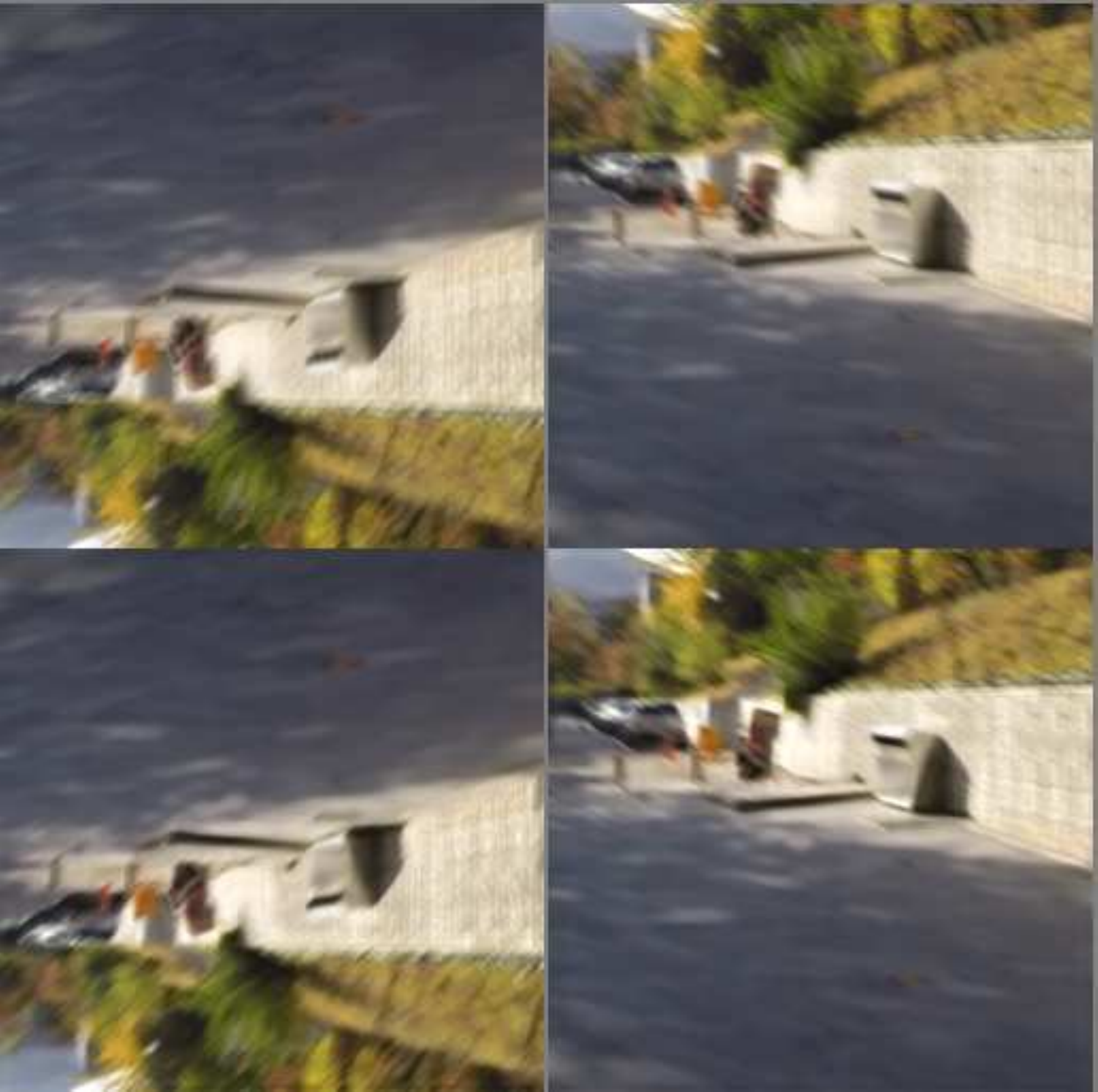}} \subfigure[$\lambda=2e^{-4}$]{\includegraphics[width=0.3\linewidth]{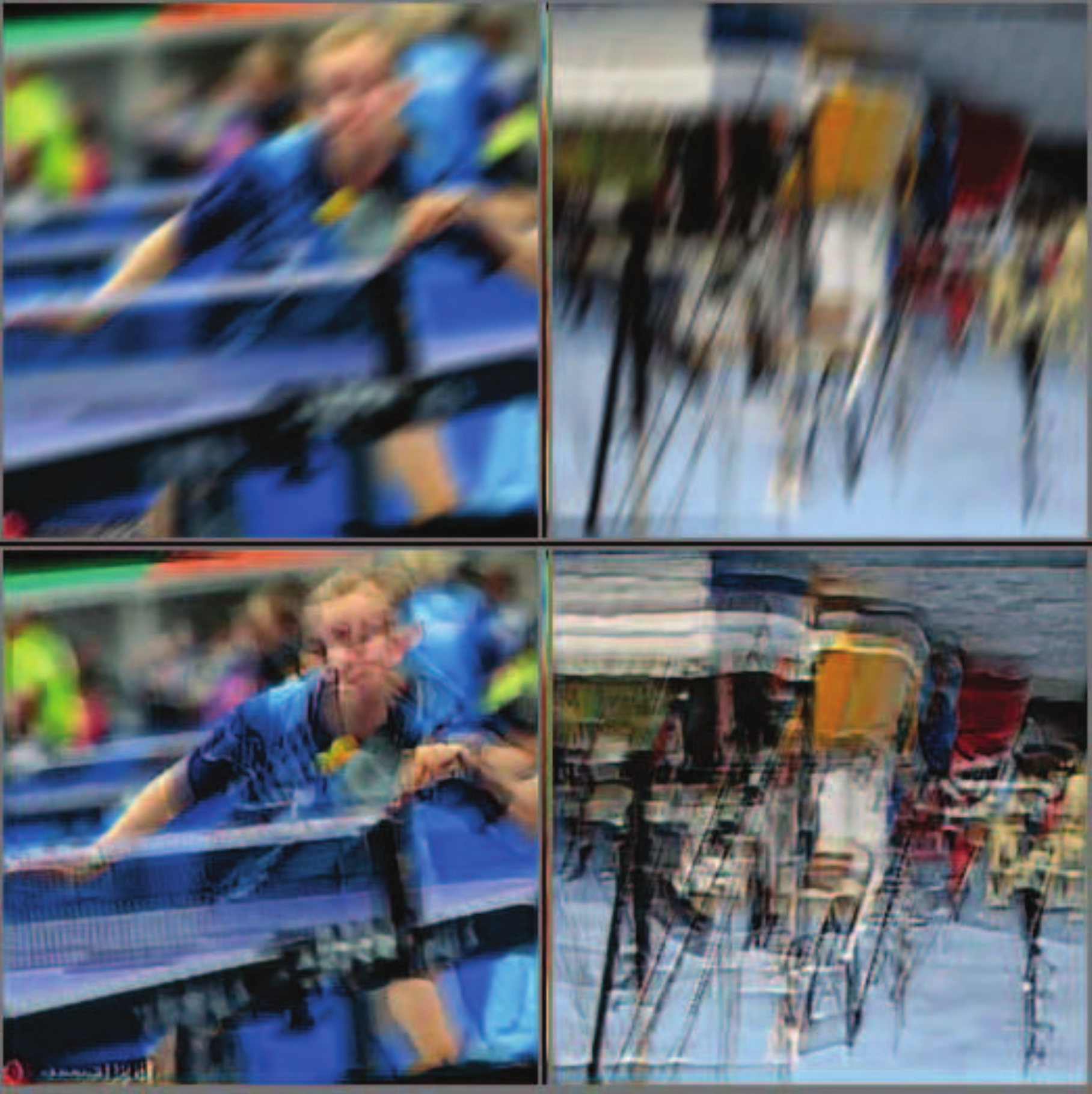}}
\subfigure[$\lambda=1e^{-2}$]{\includegraphics[width=0.3\linewidth]{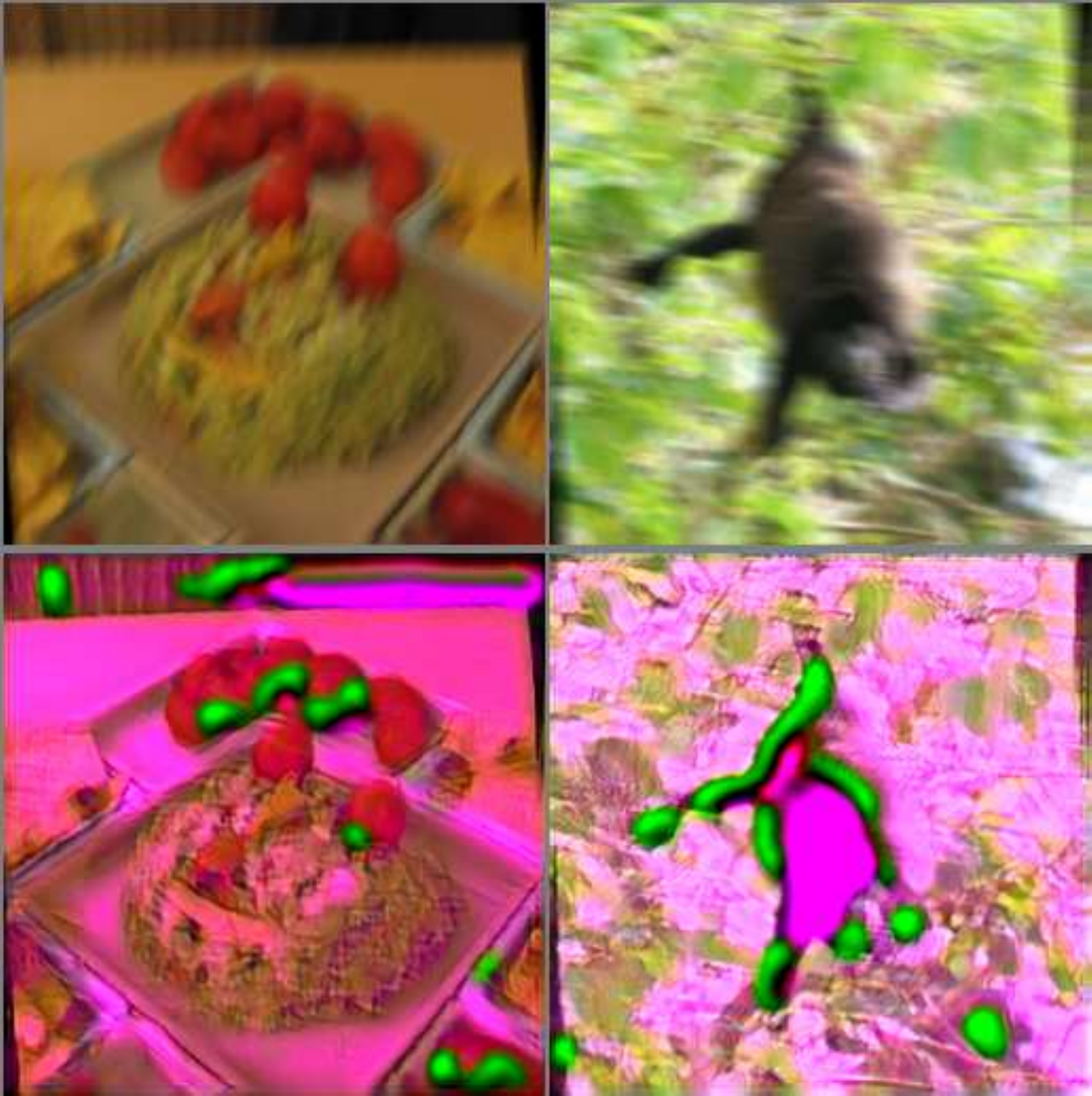}}
\\
\caption{Quality degradation of imbalanced $\lambda$. Blur input is shown in the 1st row and the model output is shown in the 2nd row. All the three models are trained to convergence with $lr=0.001$ and batch-size $=4$. It can be observed in (a) that high MSE weight setting leads to blur output; (b) shows high perceptual loss weighting would cause ghost artifacts; (c) shows much lower weighting of MSE loss could yield wrong color-histogram.
}\label{fig:imbalance_lambda}
\end{center}
\vskip -0.2in
\end{figure}

\textbf{\small Balance of Loss Functions}:
We find that the careful setting of factor $\lambda$ is required because the pixels and feature maps have different value range, and the imbalance may possibly nullify one of the metrics. Figure \ref{fig:imbalance_lambda} shows such phenomenons encountered during model adjusting.
Eventually we balance the value of $\L_c$ and $\L_p$ to be close to each other. $\lambda$ is set to $5e^{-6}$ which scale both $\L_c$ and $\L_p$ into the range $[1e^{-1}, 1e^{-2}]$.

\textbf{\small Testing settings}:
The GoPro testing-set consists of 1111 pairs of images.
The performance is calculated by averaging the evaluation values of all the testing samples.
During the training phase, we found that the inference process of network $\G$ would usually converge after 2-4 iterations. Therefore we set the iteration number to 3 as stop criterion in the testing phase, in order to evaluate the average performance under a unified setting.

\subsection{Quantitative Evaluations}
Excluding the methods that involve much more data augmentation techniques, or extended training-data sources, we compare the results with those of the state-of-the-art methods including optimization based methods: TV-$l_1$ model aided by motion flow estimation \cite{hyun2014segmentation} and $L_0$ sparsity prior based model by \cite{xu2013unnatural}; learning based methods: MBMF by \cite{gong2017motion} and CNN model predicting the distribution of motion blur by \cite{sun2015learning}.

Figure \ref{fig:examples} shows some example outputs of our method.
It can be seen that our method can deal with heterogeneous motion blur as well as homogeneous motion blur, and the sharp edges and texture details are properly recovered.
The average PSNR(dB) and SSIM results of different approaches on the GoPro dataset are shown in Table \ref{tab:comaprison}. For fairness, we use the well-recognized evaluation results on GoPro dataset collected from \cite{tao2018srndeblur,nah2017deep,DeblurGAN}.  As one can see, our method yields the best PSNR(dB) and competitive SSIM.


\begin{table}[h!]
\caption{Quantitative evaluations.
Our method performs favorably against the following compared deblurring approaches in PSNR(dB),
and is comparable to the method by \cite{gong2017motion,xu2013unnatural} in SSIM.}
\label{tab:comaprison}
\vskip 0.15in
\begin{center}
\begin{small}
\begin{sc}
\begin{tabular}{lcc}
\toprule
Metric&PSNR(dB)&SSIM\\
\midrule
\cite{sun2015learning}              &24.64 & 0.84\\ 

\cite{xu2013unnatural}              &25.18 & 0.89\\  

\cite{hyun2014segmentation}         &23.64 & 0.82\\


\cite{gong2017motion}               & 27.19 & \bf 0.90 \\ 



Ours&\bfseries 28.06 & 0.85\\  
\bottomrule
\end{tabular}
\end{sc}
\end{small}
\end{center}
\vskip -0.1in
\end{table}

\subsubsection{Runtime and Model Size}


The model size of our method is significantly smaller than many of the well-known approaches, as shown in Table \ref{tab:comparison_model_size}

\begin{table}[h!]
\caption{Comparison on Model Sizes.}
\label{tab:comparison_model_size}
\vskip 0.15in
\begin{center}
\begin{small}
\begin{sc}
\begin{tabular}{lccccc}
\toprule
Part&Size (MB)\\

\midrule

\cite{sun2015learning}&54.1\\

\cite{nah2017deep}&303.6\\

\cite{gong2017motion}&41.2\\

\cite{zhang2018dynamic}&37.1\\

Ours ($\F$) & \bfseries 1.78\\

Ours ($\G$) & \bfseries 0.46\\

Ours (Total) & \bfseries 2.24\\

\bottomrule
\end{tabular}
\end{sc}
\end{small}
\end{center}
\vskip -0.1in
\end{table}

We evaluate our method on three GPU models with different target platforms. The edition of the GTX1080Ti is for deep learning servers and the GTX1080 is for personal computers while the GT750M is for portable devices.
The results of the time-efficiency evaluation is shown in Table \ref{tab:runtime}.
The results of runtime comparison with other methods is shown in Table \ref{tab:comparison_runtime}.
Our method has the lowest inference time consumption.

\begin{table}[h!]
\caption{Run Time Analysis in Milliseconds.}
\label{tab:runtime}
\vskip 0.15in
\begin{center}
\begin{small}
\begin{sc}
\begin{tabular}{lcccc}
\toprule
Device & Year & Max & Min & Mean \\
\midrule
Gtx 1080 Ti & 2017 & 26.87 & 10.91 & 16.94 \\
Gtx 1080  & 2016 & 28.52 & 15.83 & 17.30 \\
GT 750 M  & 2013 & 106.17 & 35.97 & 40.66 \\
\bottomrule
\end{tabular}
\end{sc}
\end{small}
\end{center}
\vskip -0.1in
\end{table}

\begin{figure}[ t!]
\vskip 0.2in
\begin{center}

\includegraphics[width=0.7\linewidth]{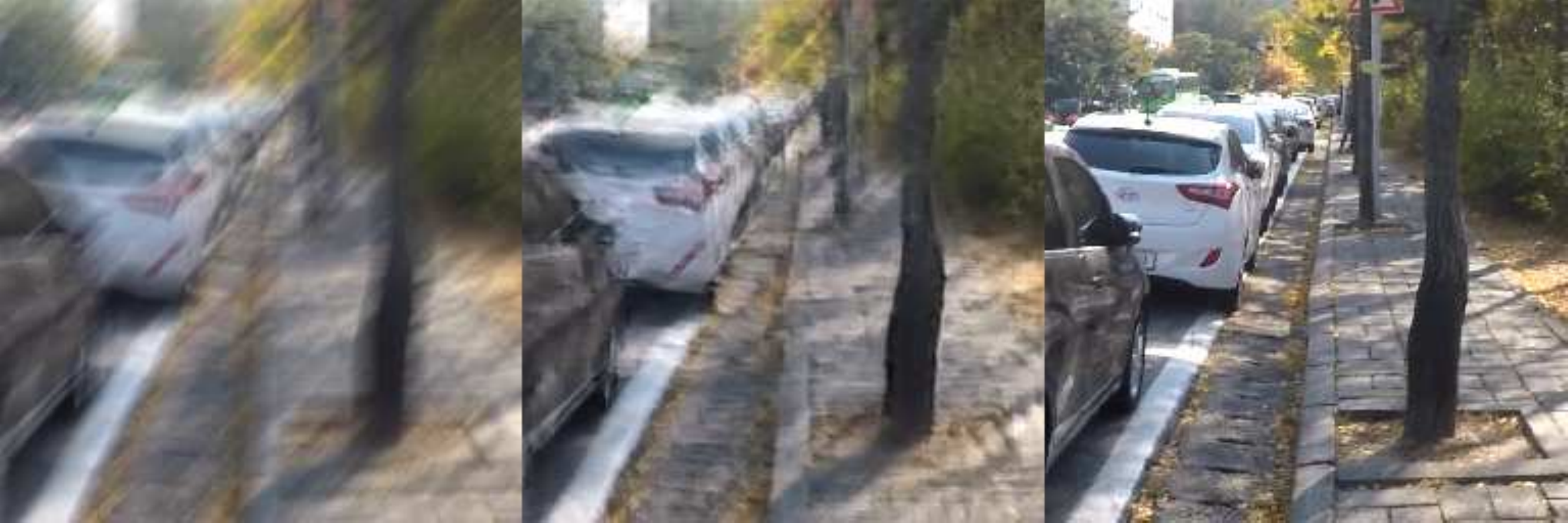}
\includegraphics[width=0.7\linewidth]{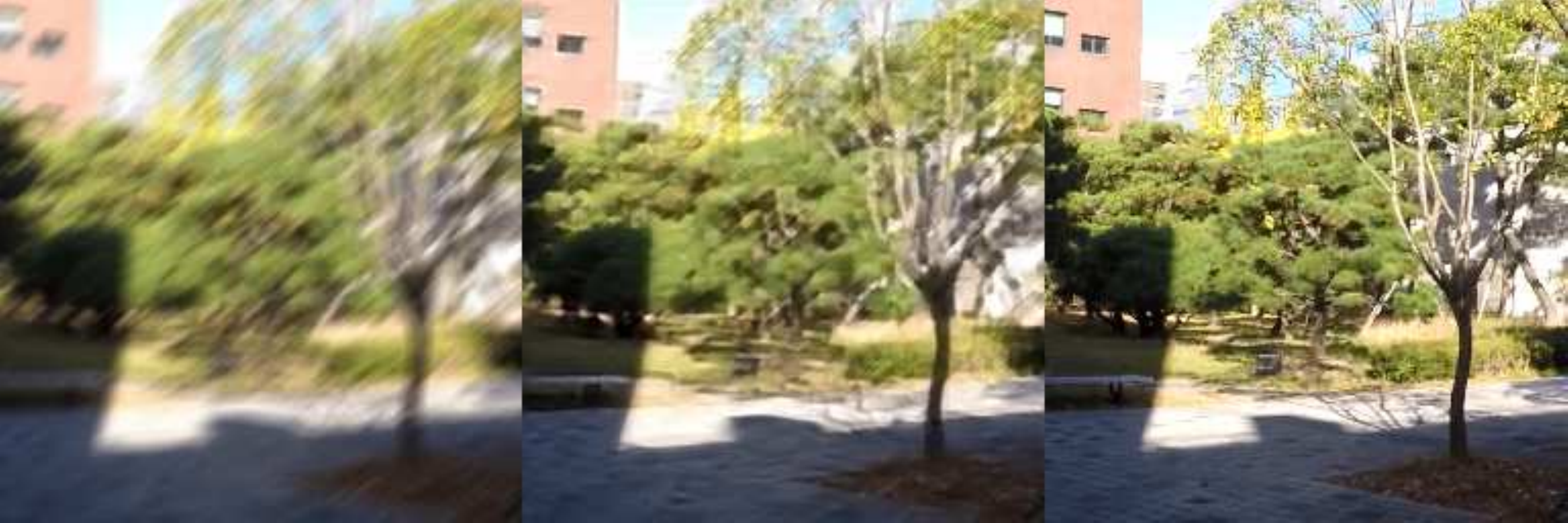}

\caption{Deconvolution results of hard examples. }\label{fig:hard_examples}
\end{center}
\vskip -0.2in
\end{figure}

\begin{table}[h!]
\caption{Runtime Comparison on Single $256\times 256$ Image.} 
\label{tab:comparison_runtime}
\vskip 0.15in
\begin{center}
\begin{small}
\begin{sc}
\begin{tabular}{lc}
\toprule
Method & Time \\
\midrule

\cite{krishnan2011blind}    &  24.23 s    \\  
\cite{levin2011efficient}   &  117.06 s   \\  
\cite{pan2016blind}         &  134.31 s   \\  
\cite{yan2017image}         &  264.78 s   \\  
\cite{li2018learning}       &  109.27 s   \\  
\cite{sun2015learning}      &  20 min     \\
\cite{xu2013unnatural}      &  1.11 s     \\  
\cite{hyun2014segmentation} &  1 h        \\
\cite{xu2010two}            &  0.80 s     \\
\cite{gong2017motion}       &  0.72 s     \\
\cite{DeblurGAN}            &  0.85 s     \\
\bfseries Ours              &  \bfseries 16.94$\sim$40.66 ms   \\

\bottomrule
\end{tabular}
\end{sc}
\end{small}
\end{center}
\vskip -0.1in
\end{table}

\subsection{Limitations}
Removing severe motion blur is a challenge of image deconvolution.
There are occasions under which our method cannot completely remove the blurring. As shown in Figure \ref{fig:hard_examples}, our network converges with remaining ghost artifacts. This is usually because network $\F$ cannot project the input image into the domain that $\G$ accepts.
Actually, in theory our framework can be applied again to further decouple the more difficult deblur task from $\F$, and that could be the future work of this method.


\section{Conclusion}
In this paper, we have proposed Golf Optimizer, a simple but novel framework to address blind image deconvolution.
The optimizer is separated into two task-dependent CNNs: one network estimates an aggressive propagation to eliminate catastrophic forgetting; while the other learns the gradient of priors from data as deep-priors to avoid the instability of prior learning.
Essentially, the optimizer can learn a good propagation for delicate correction, which is not limited to image deconvolution problem.
We have shown that training of iterative network with residual structure can asymptotically learn the gradient, and residual CNN is a reasonable architecture for the optimizer.
The network is trained and evaluated on a challenging dataset GoPro, yielding competitive performance of deconvolution.
Also, the advantages of the lightweight and runtime efficiency of the model enable it to be easily applied on portable platforms.

%
%

\nocite{langley00}

{\small
\bibliographystyle{ieee}
\bibliography{egpaper_final}
}

\section*{Appendix}
\subsection*{Gradient problems of priors}
To better understand the results of Figure \ref{fig:GradPriorExp} in our manuscript, we provide the details of experimental settings and further analysis.

We build on a simple setting to experimentally validate the advantage of gradient learning for prior .
Suppose we are to restore image $u$ consisting of only $2$ pixels: $u=(u_1,u_2)$.
An image is said to be sharp if $u_1+u_2>0$, otherwise blur.
The ground truth $\tilde{u}$ of a blur image $u$ is defined as the symmetric point w.r.t line $u_1+u_ 2=0$.

\textbf{Discriminative Prior with MLP:}

We train an MLP classifier $p$ to distinguish blurry/sharp images with $p(u) = 1$ if $u$ is blurry and $p(u) = 0$ if $u$ is sharp.
The architecture of MLP is given in Table \ref{tab:arch_MLP}.
\begin{table}[h!]
\caption{Network architecture of MLP}
\label{tab:arch_MLP}
\vskip 0.15in
\begin{center}
\begin{small}
\begin{sc}
\begin{tabular}{lcc}
\toprule
 Stage & Layer & params\\
\midrule
1   &         Linear&     12  \\
2   &           ReLU&     0   \\
3   &         Linear&     40   \\
4   &           ReLU&      0   \\
5   &         Linear&     36   \\
6   &           ReLU&     0    \\
7   &         Linear&     8    \\
8   &        Softmax&     0    \\
\midrule
\multicolumn{2}{c}{total parameters:} & 96 \\
\bottomrule
\end{tabular}
\end{sc}
\end{small}
\end{center}
\vskip -0.1in
\end{table}

The hardware and platforms for training networks are same to those for training Golf Optimizer in our manuscript. The dataset used also consists of image pairs like $(u,\tilde{u})$ in which $u_1 = -\tilde{u}_2, u_2 = -\tilde{u}_1$.
We randomly generate 16,000 samples for training.
We train the discriminative MLP network with cross-entropy loss.
Adam optimizer is adopted to train the network, with $\beta_1=0.9$, $\beta_2 = 0.999$ and a mini-batch of size $4$.
The network is trained for 4,000 iterations with learning rate $=10^{-3}$.
After the training completes, we evaluate the classification performance of discriminative MLP on 1,000 randomly sampled testing data, and the trained network achieves $99\%$ classification accuracy.

\begin{figure}[h!]
\vskip 0.2in
\begin{center}
\centerline{\includegraphics[width=1.0\linewidth]{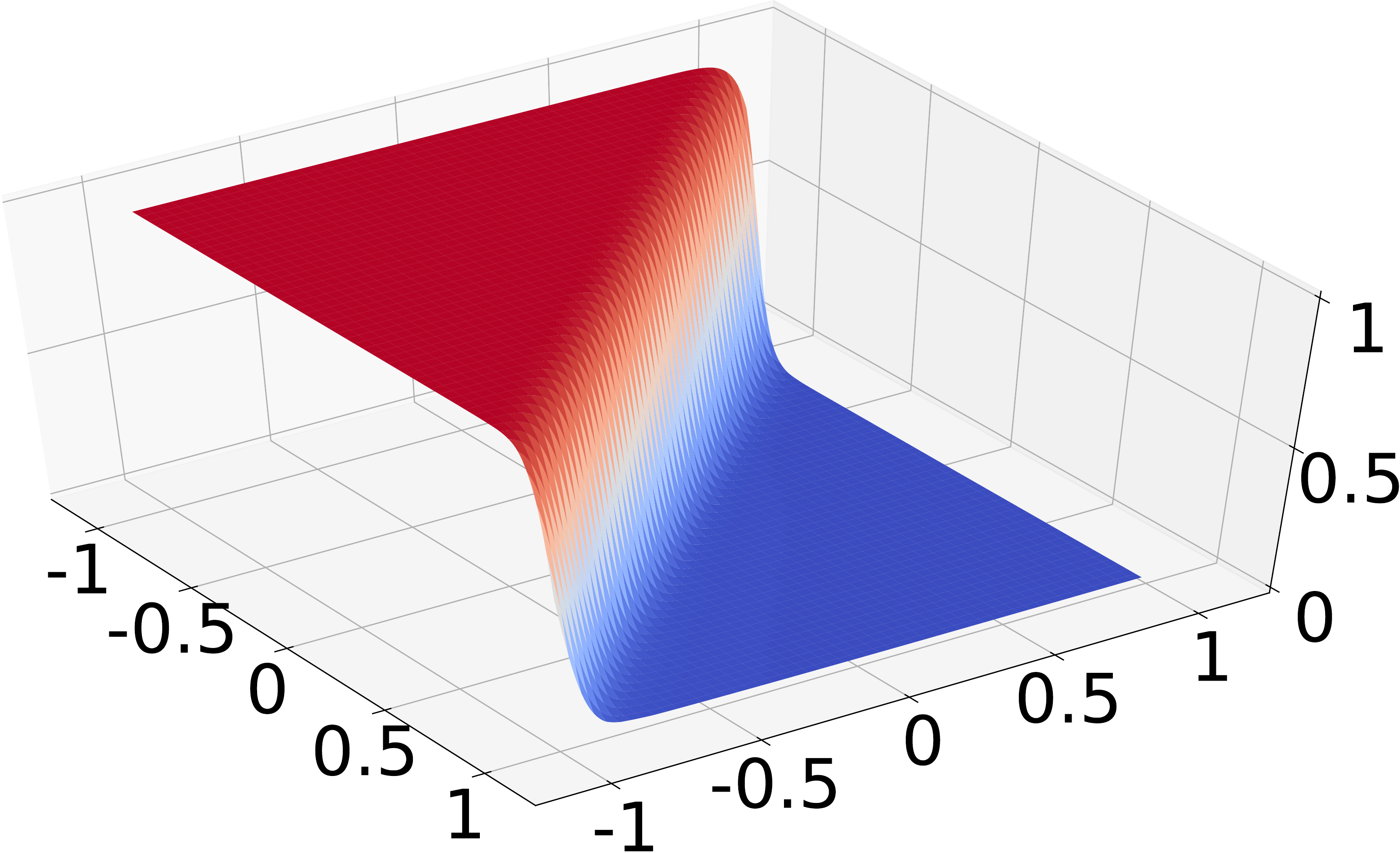}}
\caption{Instable gradient of discriminative prior.}\label{fig:GradPriorExmp1}
\end{center}
\vskip -0.2in
\end{figure}

To simulate the estimation of image with the discriminative prior, we plot the output of MLP $p(u)$ in the region  $\{(u_1,u_2) | -1\leq u_1 \leq 1, -1 \leq u_2 \leq 1\}$.
Figure \ref{fig:GradPriorExmp1} which is same as Figure \ref{fig:dis_prior} in our manuscript, shows that when using the discriminative network $p$ as prior to deblur image $u$, the response of $p$ w.r.t $u$ is $1$ with large probability in the red flat areas.
This discriminative gives high response of blurry image but zero gradient w.r.t image, which shows that discriminative MLP network has difficulty in giving effective feedbacks to aid the estimation of sharp images.
This phenomenon of gradient vanishing mostly lies on the improper evaluation metric on prior.
In this simple situation, when the blurry image $u$ is far away from the decision line $u_1 + u_2 =0$, the classifier prior only produces response with $1$, and the update for deblur expects a large step but gets almost $0$.
While the image $u$ is close to the line, the update for deblur excepts a small step but gets large one.

\begin{figure}[h!]
\vskip 0.2in
\begin{center}
\centerline{\includegraphics[width=0.8\linewidth]{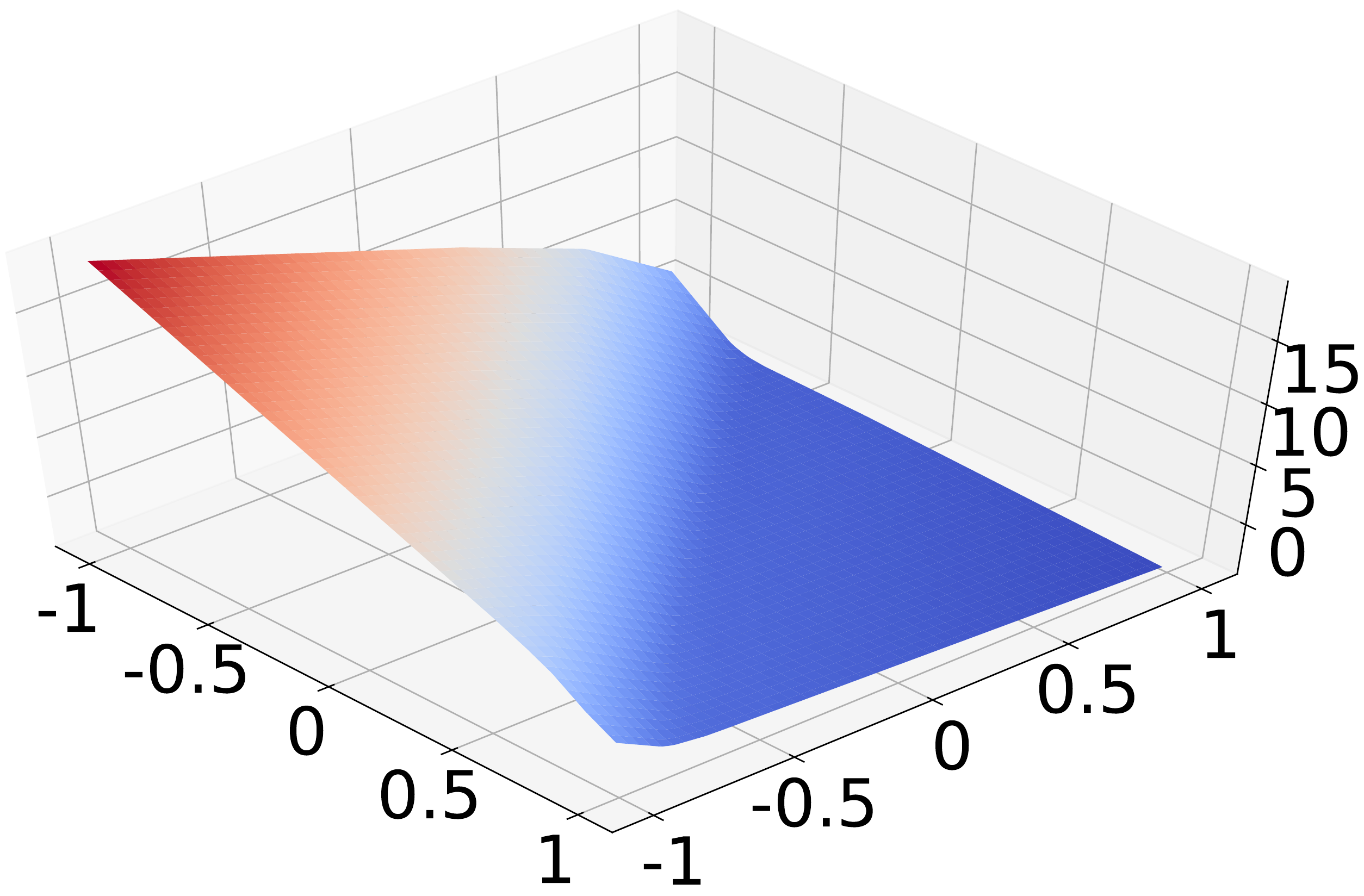}}
\caption{Plot of the logits of the discriminative prior.}\label{fig:logits}
\end{center}
\vskip -0.2in
\end{figure}

Moreover, we take only the logits from the last hidden layer, (i.e., we remove the softmax layer), of the discriminative prior for analysis, the plotted surface would takes the appearance shown in Figure \ref{fig:logits}.
As the figure shows, the gradients become large when the blurry images approach the decision line, which contradicts with the desired behaviour. Using such a prior, if we update a blur image $u$ by applying $u' = u-\nabla p$ and compute the loss by $\mathrm{MSE}=\|\tilde{u} - u'\|^2$, the loss value comes out to be 0.4849 which is quite large.

\begin{figure}[h!]
\vskip 0.2in
\begin{center}
\centerline{\includegraphics[width=0.7\linewidth]{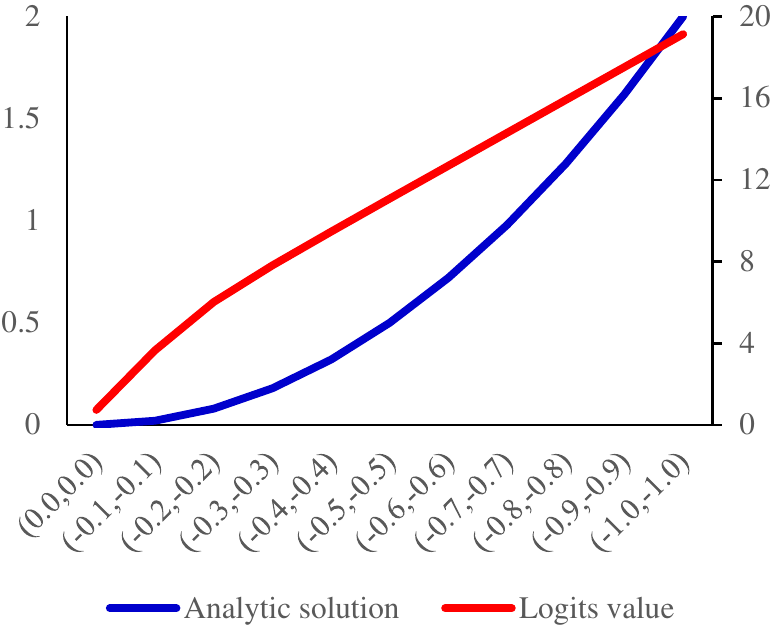}}
\caption{A sliced view of the function surface of the discriminative prior compared with that of the analytic solution. Values are sampled from the function surface. X-axis is the sampled location $(u_1,u_2)$ and z-axis corresponds to the response value of priors. We can see that the gradient of the discriminator is smaller at $(-1,-1)$ and larger at (0,0), which is not proportional to the distance between $u$ and its ground truth $\tilde{u}$. }\label{fig:grad_trend}
\end{center}
\vskip -0.2in
\end{figure}

\begin{figure}[!htp]
\vskip 0.2in
\begin{center}
\centerline{\includegraphics[width=0.8\linewidth]{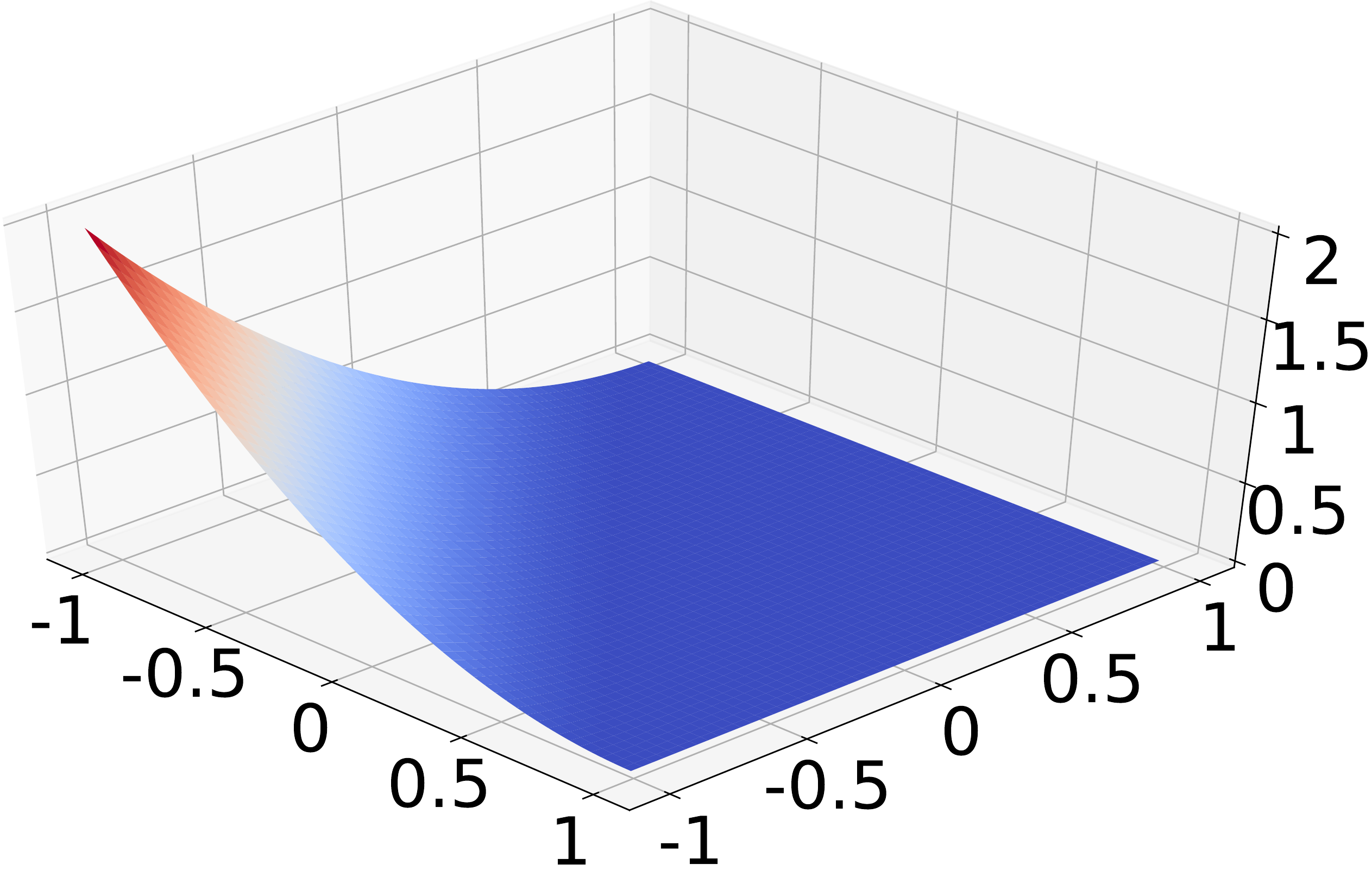}}
\caption{Analytical optimal solution.}\label{fig:optimal_p}
\end{center}
\vskip -0.2in
\end{figure}

Besides, even if the general descent trend of the discriminative prior seems to be roughly similar to the desired form, if we precisely compare the descent trend of the discriminative prior in Figure \ref{fig:logits} and the analytic solution in Figure \ref{fig:optimal_p}, it could be very obvious to see that the discriminative prior doesn't perform well, as shown in Figure \ref{fig:grad_trend}.

This experiment shows that, without careful design and improvement, prior learning with discriminative criterion could be problematic and not optimal.

\textbf{Prior learning with gradient constraint:}

In contrast to the priors designed or learnt with other targets in which the gradient is rarely constrained, we instead apply our gradient learning framework to the task.

Since we are to update the image $u$ so that it falls near its ground truth $\tilde{u}$ on the other side of the decision line $u_1 + u_2 = 0$ , we learn the desired prior by learning the negative of its gradient: $f(u) = \tilde{u} - u = - \nabla p$. By doing this, $f$ learns both the descent direction $(\tilde{u} - u) / \| \tilde{u} - u \|$ and the optimal descent step size $\| \tilde{u} - u \|$. Formally, we train the network $f$ by minimizing $\| f(u) - (\tilde{u} - u ) \|^2$. The network architecture is shown in Table \ref{tab:arch_MLP_GradPrior}.

\begin{table}[h!]
\caption{Architecture of our method to learn the gradient of prior.}
\label{tab:arch_MLP_GradPrior}
\vskip 0.15in
\begin{center}
\begin{small}
\begin{sc}
\begin{tabular}{lcc}
\toprule
 Stage & Layer & params\\
\midrule
1   &         Linear&     12  \\
2   &           ReLU&     0   \\
3   &         Linear&     40   \\
4   &           ReLU&      0   \\
5   &         Linear&     36   \\
6   &           ReLU&     0    \\
7   &         Linear&     8    \\
\midrule
\multicolumn{2}{c}{total parameters:} & 96 \\
\bottomrule
\end{tabular}
\end{sc}
\end{small}
\end{center}
\vskip -0.1in
\end{table}

\begin{figure}[h!]
\vskip 0.2in
\begin{center}
\centerline{\includegraphics[width=0.8\linewidth]{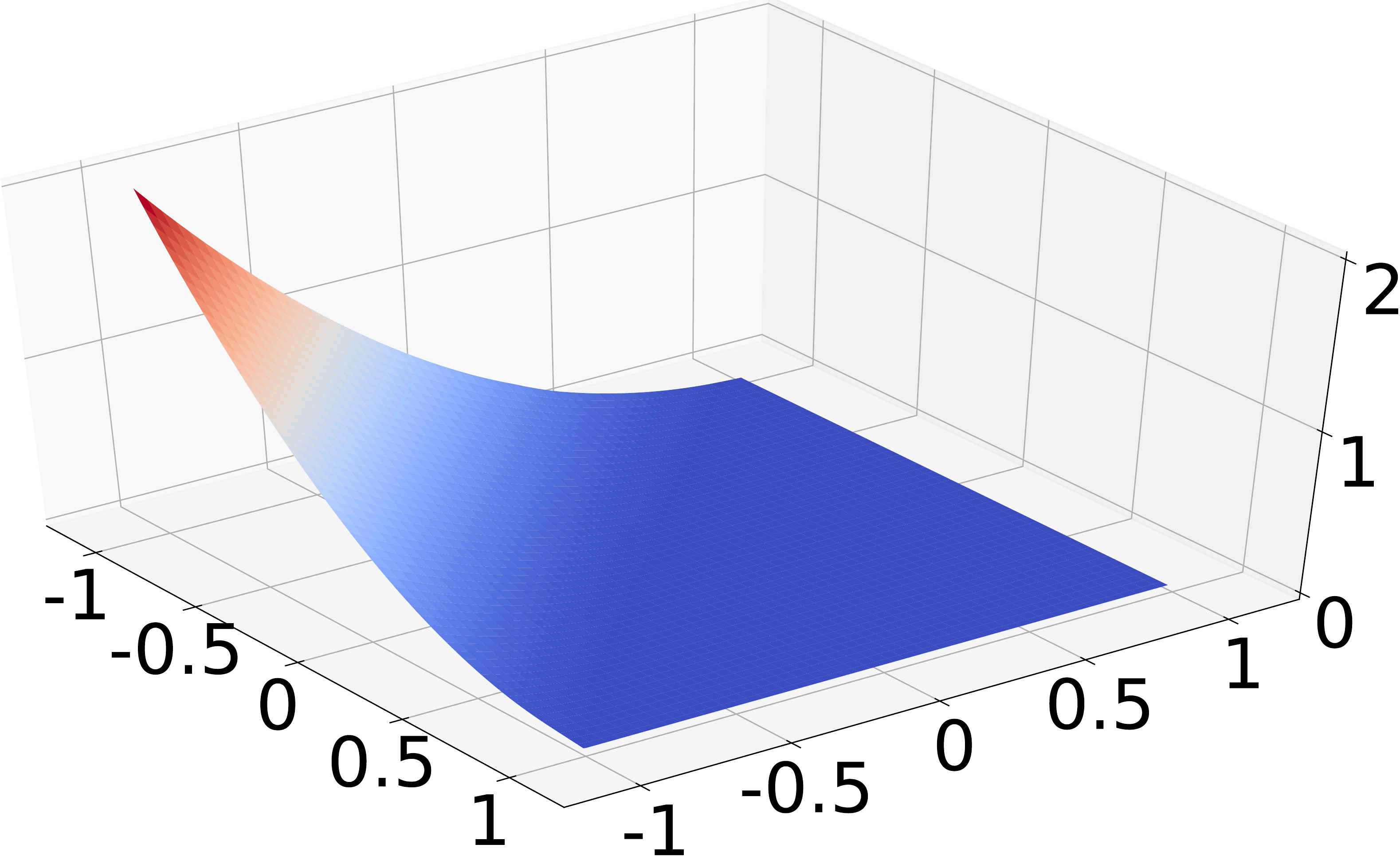}}
\caption{Prior with constrained gradient.}\label{fig:GradPriorExmp2}
\end{center}
\vskip -0.2in
\end{figure}

For better interpretability, we construct the prior $p$ from its negative gradient $f$ which is what we have learnt. We use numerical integration to do the reconstruction and the result is shown in \ref{fig:GradPriorExmp2}.
The negative of gradient of the plotted surface is the output of $f$.

Compared with the discriminative prior whose testing loss is $0.4849$ in terms of MSE, the prior that learns the gradient of prior achieves MSE $=2e^{-3}$ on 1000 randomly sampled testing pairs.
Similar to the previous experiment, this result of MSE is computed as follow.
Given a pairs of testing sample $(u,\tilde{u})$, the update is performed by gradient descent method as $u' = u + f(u)$, and the MSE is computed by $\|\tilde{u} - u'\|^2$.

Furthermore, we construct the optimal solution as
\begin{equation} \nonumber
  p^*(u) = \frac{1}{2} (u_1 + u_2)^2,
\end{equation}
and the plotted surface is shown in Figure \ref{fig:optimal_p}.
The optimal solution $p^*$ is constructed with the objective that its negative of gradient is equivalent to the difference of $\tilde{u}$ and $u$, that is $- \nabla_u p^* = \tilde{u} - u$.
According to our setting, with $u=(u_1, u_2)$, we have $\tilde{u} = (-u_2, -u_1)$.
The derivative of $p^*$ w.r.t $u_1$ and $u_2$ is
\begin{equation} \nonumber
  \frac{\partial p^*}{\partial u_1} = \frac{\partial p^*}{\partial u_2} = u_1 + u_2.
\end{equation}
The update of $u$ with gradient descent is performed as
\begin{equation} \nonumber
  \begin{split}
    u' &= u - \nabla_u p^* = (u_1, u_2) - (u_1+u_2, u_1+u_2) \\
    &= (-u_2, -u_1) = \tilde{u}.
  \end{split}
\end{equation}
Hence, the construction of optimal solution is reasonable.
From Figure \ref{fig:optimal_p}, it can be seen that it is coincident to the prior obtained by learning the gradient. In fact, formally the MSE between this analytical solution and the prior is the same as the MSE between the prior and ground truth data, which is also $2e^{-3}$.

\newpage
\section*{More experimental results}
 Shown in left is the input blur image, while the output and ground-truth are in middle and right respectively.

\begin{figure}[!h]
\vskip 0.2in
\begin{center}
\subfigure[PSNR(dB):35.73, SSIM:0.96]{\includegraphics[width=\linewidth]{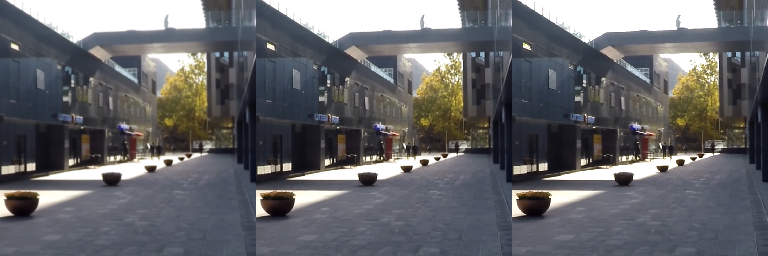}\label{fig:gopro_platta}}
\subfigure[PSNR(dB):30.16, SSIM:0.88]{\includegraphics[width=\linewidth]{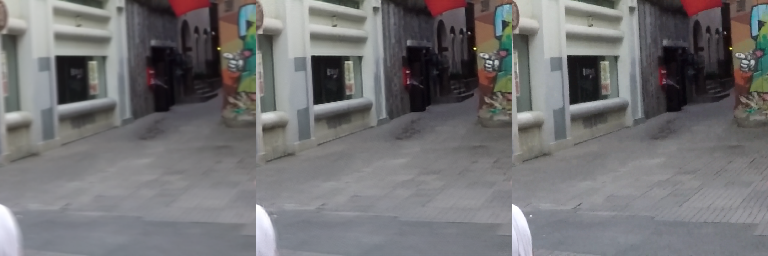}\label{fig:gopro_corner}}
\subfigure[PSNR(dB):30.43, SSIM:0.91]{\includegraphics[width=\linewidth]{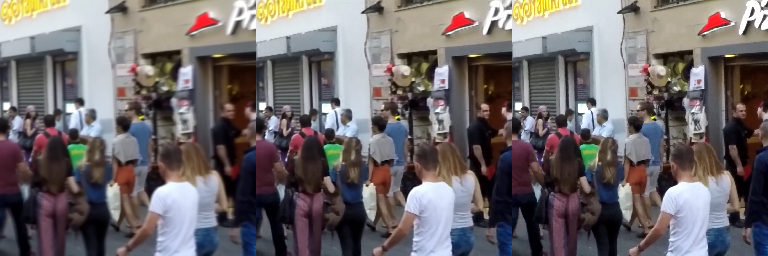}\label{fig:gopro_pizzahut}}
\subfigure[PSNR(dB):28.66, SSIM:0.83]{\includegraphics[width=\linewidth]{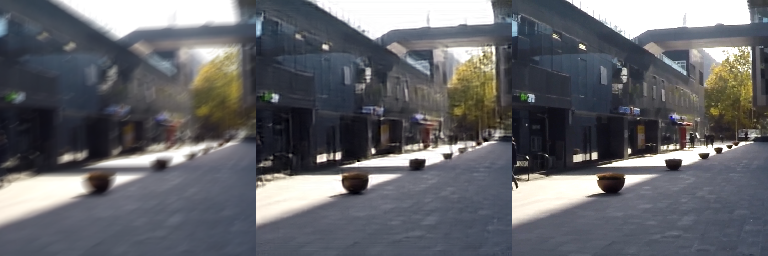}\label{fig:gopro_platta_worse}}
\subfigure[PSNR(dB):28.42, SSIM:0.86]{\includegraphics[width=\linewidth]{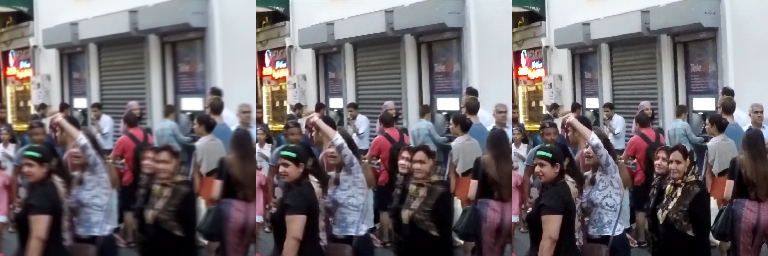}\label{fig:gopro_india}}
\caption{Example results (Part-I)}\label{fig:examples1}
\end{center}
\vskip -0.2in
\end{figure}

\begin{figure}[!htp]
\vskip 0.2in
\begin{center}
\subfigure[PSNR(dB):27.03, SSIM:0.80]{\includegraphics[width=\linewidth]{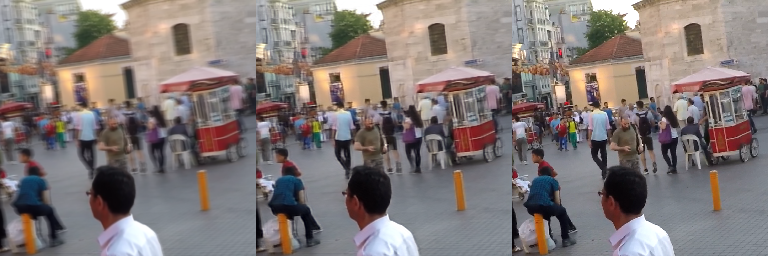}\label{fig:gopro_castle}}
\subfigure[PSNR(dB):27.82, SSIM:0.82]{\includegraphics[width=\linewidth]{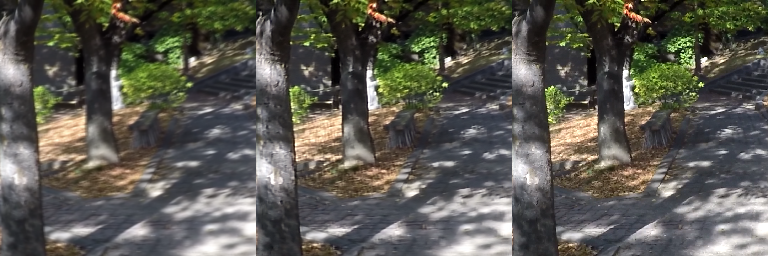}\label{fig:gopro_trees}}
%
\subfigure[PSNR(dB):26.06, SSIM:0.69]{\includegraphics[width=\linewidth]{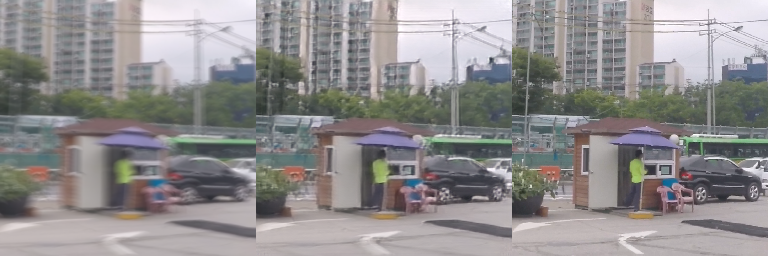}\label{fig:gopro_toll}}
\subfigure[PSNR(dB):25.32, SSIM:0.74]{\includegraphics[width=\linewidth]{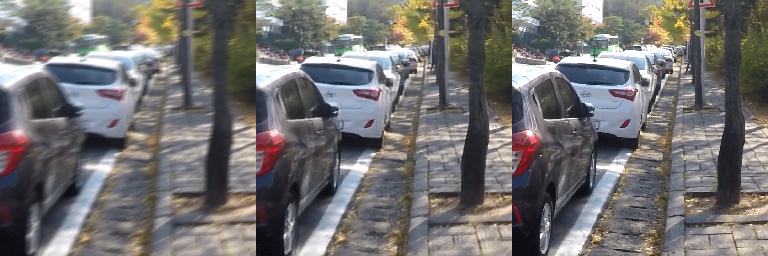}\label{fig:gopro_car}}
\subfigure[PSNR(dB):25.83, SSIM:0.76]{\includegraphics[width=\linewidth]{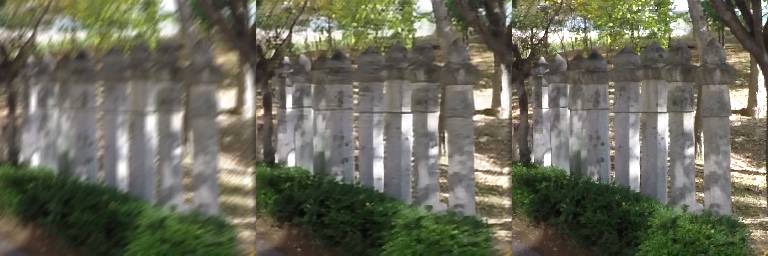}\label{fig:gopro_sticks}}
\subfigure[PSNR(dB):31.69, SSIM:0.90]{\includegraphics[width=\linewidth]{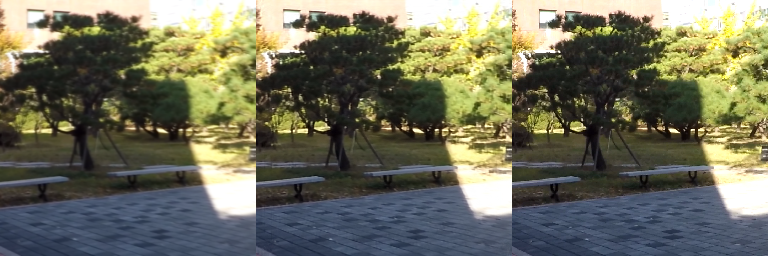}\label{fig:gopro_p29}}

\caption{Example results (Part-II)}\label{fig:examples2}
\end{center}
\vskip -0.2in
\end{figure}

\end{document}